\documentclass[]{memtensor}

\usepackage[toc,page,header]{appendix}
\usepackage[utf8]{inputenc} 
\usepackage[T1]{fontenc}    
\usepackage{hyperref}       
\usepackage{url}            
\usepackage{booktabs}       
\usepackage{amsfonts}       
\usepackage{nicefrac}       
\usepackage{microtype}      
\usepackage{xcolor}         
\usepackage{amsmath} 
\usepackage{etoolbox}
\usepackage{lipsum}  
\usepackage{minitoc}
\usepackage[table]{xcolor}  
\usepackage{tablefootnote}  
\usepackage{fancyhdr}

\usepackage{graphicx}
\usepackage{multirow}
\usepackage[flushleft]{threeparttable}  
\usepackage{titletoc}
\usepackage{siunitx} 
\usepackage{colortbl}
\usepackage{tcolorbox}
\usepackage{listings}
\usepackage{minted}
\usepackage{makecell}
\usepackage{marvosym}
\usepackage{lmodern}
\usepackage{fix-cm}
\usepackage{tikz}

\definecolor{softgreen}{RGB}{120,200,200}
\definecolor{softbeige}{RGB}{255,253,250}
\definecolor{softred}{RGB}{220,120,120}

\newcommand{\lightmidrule}{\arrayrulecolor{gray!60}\specialrule{0.5pt}{1pt}{1pt}\arrayrulecolor{black}}

\title{HaluMem: Evaluating Hallucinations in Memory Systems of Agents}

\author[1]{Ding Chen*}
\author[2]{Simin Niu*}
\author[2]{Kehang Li}
\author[2]{Peng Liu}
\author[]{Xiangping Zheng$^3$\Letter}
\author[2]{Bo Tang}
\author[1]{Xinchi Li}
\author[2]{Feiyu Xiong}
\author[]{Zhiyu Li$^2$\Letter}

\affiliation[1]{China Telecom Research Institute}
\affiliation[2]{MemTensor (Shanghai) Technology}
\affiliation[3]{Harbin Engineering University}

\abstract{
\vspace{-0.75em}
Memory systems are key components that enable AI systems such as LLMs and AI agents to achieve long-term learning and sustained interaction. However, during memory storage and retrieval, these systems frequently exhibit memory hallucinations, including fabrication, errors, conflicts, and omissions. Existing evaluations of memory hallucinations are primarily end-to-end question answering, which makes it difficult to localize the operational stage within the memory system where hallucinations arise. To address this, we introduce the Hallucination in Memory Benchmark (HaluMem), the first operation level hallucination evaluation benchmark tailored to memory systems. HaluMem defines three evaluation tasks (memory extraction, memory updating, and memory question answering) to comprehensively reveal hallucination behaviors across different operational stages of interaction. To support evaluation, we construct user-centric, multi-turn human-AI interaction datasets, HaluMem-Medium and HaluMem-Long. Both include about 15k memory points and 3.5k multi-type questions. The average dialogue length per user reaches 1.5k and 2.6k turns, with context lengths exceeding 1M tokens, enabling evaluation of hallucinations across different context scales and task complexities. Empirical studies based on HaluMem show that existing memory systems tend to generate and accumulate hallucinations during the extraction and updating stages, which subsequently propagate errors to the question answering stage. Future research should focus on developing interpretable and constrained memory operation mechanisms that systematically suppress hallucinations and improve memory reliability. 
\vspace{-0.75em}
}

\date{\today}
\correspondence{\email{xpzheng@hrbeu.edu.cn, lizy@memtensor.cn}}
\checkdata[Author Legend]{*Co-equal primary author}
\checkdata[Code]{\url{https://github.com/MemTensor/HaluMem}}
\checkdata[Datasets]{\url{https://huggingface.co/datasets/IAAR-Shanghai/HaluMem}}

\begin{document}
\maketitle

\section{Introduction}

Each interaction between a user and an LLM may contain personalized information about the user~\cite{shi2024wildfeedback,zhao2025do}. However, such information is often forgotten once the conversation ends, making it difficult for the model to continuously understand the user, adapt to persona shifts, or generate personalized responses~\cite{liu2023thinkinmemory,zhang2024guided}. To ensure that LLMs maintain coherence and personalization in long-term interactions, it is crucial to develop a mechanism capable of recording, updating, and utilizing user information, which constitutes the core function of a memory system.

A memory system serves as the fundamental infrastructure for organizing and managing information based on the history of human–AI conversations. It extracts, structures, and continuously updates key information generated across multi-turn interactions between users and AI systems, retrieving and injecting this information into the model as needed to support personalization and long-term consistency~\cite{li2025memos_long,memobase,kang2025memory,rasmussen2025zep,supermemory}. Specifically, a memory system identifies stable user profiles, narratives, and events from dialogues and stores them as plaintext entries enriched with metadata. When new queries or tasks arise, the system retrieves and selectively integrates relevant memories based on the current intent and context, enabling the AI system to “remember and correctly utilize” user information, thereby preserving semantic coherence, behavioral consistency, and preference alignment. Representative systems such as MemOS~\cite{li2025memos_long}, Mem0~\cite{chhikara2025mem0}, Zep~\cite{rasmussen2025zep}, Supermemory~\cite{supermemory}, and Memobase~\cite{memobase} continuously record user profiles, events, and evolving preferences, supporting the creation, revision, and tracking of memories to construct a system-level memory layer with structured management capabilities.

\begin{figure*}[htp]
    \centering
    \includegraphics[width=1\linewidth]{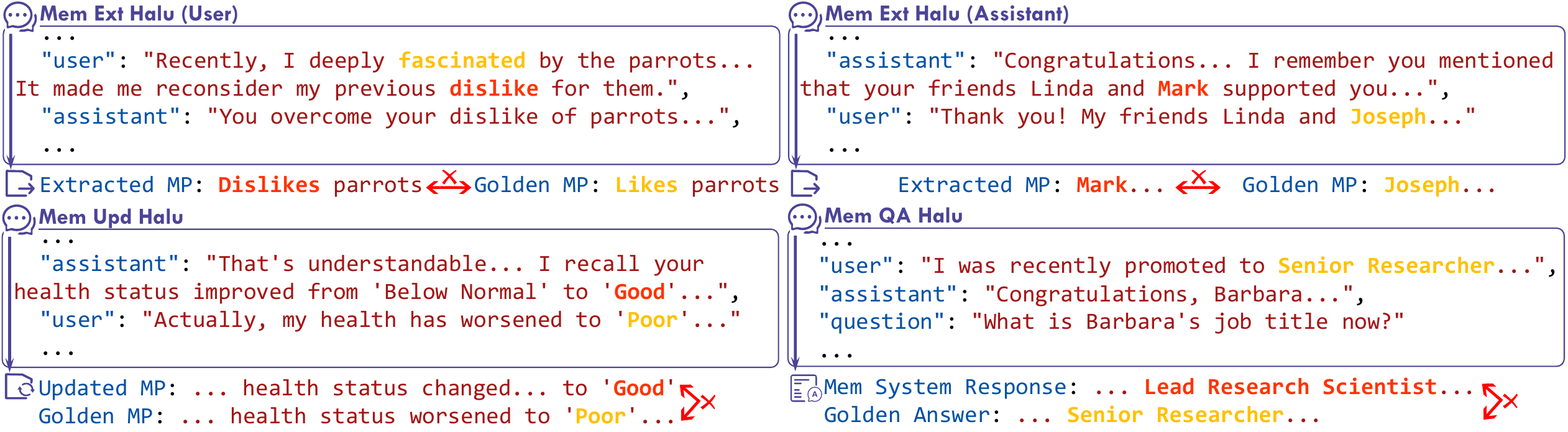}
    \caption{Examples of operation-level hallucination in a memory system.} 
    \label{fig:HaluMem_Case}
\end{figure*}

Although these systems significantly improve the organization and controllability of memory, they are commonly affected by the phenomenon of memory hallucination (Figure~\ref{fig:HaluMem_Case}), which manifests as fabricated, erroneous, conflicting, or missing information during the processes of memory extraction or updating~\cite{DBLP:journals/corr/abs-2507-18910,DBLP:conf/fllm/AgrawalKA024,math13050856}. Such issues undermine the accuracy and consistency of memory. More importantly, these upstream hallucinations are often amplified during the generation stage, further inducing generation hallucination~\cite{10.1145/3703155} and compromising the overall reliability of the system. To effectively mitigate these hallucination phenomena, it is essential to establish a systematic hallucination evaluation mechanism for memory systems. However, existing methods for evaluating hallucinations in memory systems remain limited. Most mainstream studies adopt end-to-end, question–answer-based evaluation frameworks that assess memory quality indirectly through the final output performance of AI systems, making it difficult to determine at which stage of the memory process the hallucination originates.

To address this issue, we propose the Hallucination in Memory Benchmark (HaluMem), the first operation-level hallucination evaluation benchmark for memory systems, which comprises two datasets, HaluMem‑Medium and HaluMem‑Long. HaluMem constructs an evaluation framework that encompasses three types of tasks: memory extraction, memory updating, and memory question answering, in order to comprehensively reveal the hallucination behaviors of memory systems across different operational levels of interaction.

However, achieving operation-level evaluation of hallucinations in memory systems is a nontrivial task, as it requires a multi-turn conversational dataset that can comprehensively represent the processes of memory generation, updating, and retrieval across interactions\footnote{The dataset we designed is a user-centered multi-turn human–AI conversation dataset. This design is motivated by the fact that memory systems are inherently intended to support personalized and long-term human–AI interactions. Their core functionality lies in organizing, storing, and updating user-related memories throughout continuous dialogue, which enables a more realistic evaluation of how hallucinations occur when memory systems organize and update memories around the user.}. Such a dataset is used as input to the memory system under evaluation and requires the system to identify and process memories of different operational types that accumulate throughout the conversation. To this end, we provide a multi-turn conversation dataset with precise annotations for each memory operation and its corresponding result, referred to as a memory point. By comparing the memory points produced by a memory system with the annotated ground-truth memory points, we can perform fine-grained evaluation to determine whether hallucinations occur in memory extraction, memory updating, or question answering (i.e., memory retrieval). Specifically, we measure accuracy and coverage to assess hallucinations caused by errors or fabrications in memory extraction, use consistency to evaluate hallucinations arising from errors or omissions in memory updating, and identify hallucinations in question answering that result from incorrect references or fabricated content.

Based on this design, we construct two benchmark datasets: HaluMem-Medium and HaluMem-Long. Both datasets contain approximately 15,000 memory points and more than 3,400 evaluation queries, with each user involved in over one thousand conversational turns on average. The latter further extends the average dialogue context length per user to the scale of millions of tokens, allowing examination of hallucination behaviors in ultra-long conversations.

The main contributions of this work are summarized as follows:

\begin{itemize}
    \item We propose HaluMem, the first operation-level benchmark for evaluating memory hallucinations, which overcomes the limitations of prior end-to-end evaluation methods by systematically revealing hallucination phenomena across three operational dimensions: memory extraction, memory updating, and memory-based question answering.
    \item We construct an extensive multi-turn evaluation dataset for human–AI interactions and design two user-centered benchmarks, HaluMem-Medium and HaluMem-Long, to evaluate hallucination behaviors of memory systems under different contextual scales and task complexities.
    \item Through stage-wise evaluation, we reveal the cumulative and amplifying effects of hallucinations across memory extraction, updating, and question answering, providing a new analytical perspective for understanding and mitigating hallucinations in memory systems.
\end{itemize}

\section{Related Work}

\subsection{Memory System}

Large Language Models (LLMs) and AI Agents built upon them possess implicit memory capabilities, where knowledge is primarily embedded within model parameters through large-scale pre-training and fine-tuning, thereby forming a parameterized form of long-term memory. Although such implicit memory enables LLMs to demonstrate strong knowledge recall and reasoning abilities during inference and generation, it suffers from poor manageability—the internal memory of the model cannot be explicitly accessed, updated, or deleted, and there is no mechanism for controlling its lifecycle. When encountering outdated or conflicting information, the model often struggles to revise or replace old knowledge, which can lead to memory hallucination, manifesting as the generation of erroneous, obsolete, or inconsistent content.

\begin{table}[htbp]
\caption{Comparison of Various Memory Systems.}
\centering
\small
\resizebox{1.0\linewidth}{!}{
\begin{tabular}{@{}llccc@{}}
\toprule
\textbf{Method} & \textbf{Memory Type} & \textbf{Memory Operation} & \textbf{Manageability} & \textbf{Graph Support} \\ 
\midrule



Supermemory~\cite{supermemory} & Plain Text (with Metadata) & CUDE & fair–Excellent & Yes \\

Memobase~\cite{memobase} & Plain Text (with Metadata) & CUD & Excellent & No \\

Zep~\cite{rasmussen2025zep} & Plain Text (with Metadata) & CUD & fair–Excellent & Yes \\

Mem0~\cite{chhikara2025mem0} & Plain Text (with Metadata) & CUD & fair–Excellent & Yes \\

MemOS~\cite{li2025memos_long} & Parameter; Activation; Plain Text (with Metadata) & CUD & fair–Excellent & Yes \\

\bottomrule
\end{tabular}
}
\label{tab:memory-systems}
\end{table}

Various forms of external memory modules have been proposed to address the limitations of parameterized memory. Early external memory mechanisms were primarily represented by Retrieval-Augmented Generation (RAG). In particular, RAG~\cite{lewis2020rag} introduces an external plaintext knowledge retrieval mechanism. Before generation, relevant documents are retrieved from a vector database and incorporated into the model’s input, enabling controllable and updatable external memory. This approach offers high manageability, and because the external memory is transparent and editable, it exhibits a relatively low degree of memory hallucination. However, traditional RAG systems primarily rely on text-based memory structures and lack explicit modeling of inter-entity relationships, meaning they do not support graph structures. Consequently, they remain limited in handling complex knowledge reasoning and maintaining long-term consistency. Building upon this, GraphRAG~\cite{edge2025graphrag} further integrates a knowledge graph structure, organizing and retrieving knowledge in the form of entity–relation pairs. By leveraging graph indexing and multi-hop path retrieval, GraphRAG significantly enhances the representational capacity and retrieval accuracy of structured knowledge, leading to improved performance in relational reasoning. Nevertheless, the construction and maintenance of graph structures entail high costs, and synchronization during updates introduces additional complexity. As a result, GraphRAG demonstrates moderate manageability, inferior to that of RAG, and may introduce additional memory hallucinations due to inconsistent updates among graph nodes or edges.

With the growing demand for AI systems capable of personalized interaction and long-term learning, researchers have begun exploring memory systems that possess genuine long-term maintainability and operational controllability, as summarized in Table~\ref{tab:memory-systems}. 
Supermemory~\cite{supermemory} provides long-term memory for language models by combining document retrieval and user-specific memory. It integrates both retrieval-augmented generation and agent memory, using a contextual graph that captures temporal, relational, and personal information, enabling consistent and personalized responses across interactions.
Memobase~\cite{memobase} focuses on user-level long-term memory by recording preferences and interaction histories in a plaintext structure. During interactions, it dynamically generates context snippets from user profiles and recent events to enable personalized recall, though some risk of hallucination may occur during memory extraction.
Zep~\cite{rasmussen2025zep} introduces a context engineering framework that integrates agent memory, Graph RAG, and context assembly capabilities, with its core component Graphiti enabling temporally-aware synthesis of conversational and business data for personalized long-term context.
Mem0~\cite{chhikara2025mem0} employs a metadata-enriched plaintext storage format that supports comprehensive memory operations (Create/Extract, Update, Delete, and Expand/Enrich), incorporating conflict detection and memory merging to ensure consistency and traceability.
MemOS~\cite{li2025memos_long} attempts to abstract memory as a system-level resource by unifying the management of three types of memory: parametric memory, activation memory, and explicit (plaintext) memory. Through lifecycle control, version management, and migration mechanisms, MemOS enables cross-model and cross-session memory sharing and integration. 
However, while graph structures enhance the expressiveness of memory representation, they also increase management complexity and make the system more prone to hallucination.

\subsection{Evaluation Hallucinations in Memory Systems}

\begin{table*}[htbp]
    \centering
    \caption{HaluMem vs. Existing End-to-End Benchmarks for Memory System Evaluation}
    \label{tab:memory-benchmarks-transposed}
    \resizebox{\linewidth}{!}{
    \begin{threeparttable}
        \begin{tabular}{lccccc}
        \toprule
        \textbf{Feature} & \textbf{HaluMem} & \textbf{PersonaMem} & \textbf{LOCOMO} & \textbf{LongMemEval} & \textbf{PrefEval} \\
        \lightmidrule
        Evaluation Granularity & Operation-level & End-to-end & End-to-end & End-to-end & End-to-end \\
        \lightmidrule
        Evaluation Timing & After each session & After all sessions & After all sessions & After all sessions & After all sessions \\
        \lightmidrule
        Evaluation Tasks & \makecell[c]{Memory Extraction, \\Memory Updating, \\Memory QA} & Multiple Choice & \makecell[c]{QA, \\Summarization, \\Generation} & \makecell[c]{QA,\\Memory Recall} & \makecell[c]{Generation,\\Classification} \\
        \lightmidrule
        Memory Type & \makecell[c]{Persona,\\Event,\\Relationship} & Persona & \makecell[c]{Persona,\\Event} & \makecell[c]{Persona,\\Event} & Persona \\
        \lightmidrule
        Memory Update & Yes & Yes & No & Yes & Yes \\
        \lightmidrule
        Conversation Time Span & 10$\sim$20 years & Several years* & Several months & $\sim$ 2.5 years & - \\
        \lightmidrule
        Avg Length / Session & 8.3k tokens & 6k tokens & 477 tokens & 3k tokens & - \\
        \lightmidrule
        Max Context Length & 1M tokens & 1M tokens & 9k tokens & 1.5M tokens & 100k tokens \\
        \lightmidrule
        Question Num & 3,467 & $\sim$ 6,000 & 7,512 & 500 & 3,000 \\
        \bottomrule
        \end{tabular}
        \begin{tablenotes}
        \footnotesize
        \item *~“Several years” for PersonaMem is inferred from the paper and dataset, not explicitly labeled.
        \end{tablenotes}
    \end{threeparttable}
    }
\end{table*}

Hallucinations in memory systems can be divided into two types: memory hallucinations and generation hallucinations. The former refers to inconsistencies or errors that occur during the processes of storing, updating, or retrieving information within a memory system, such as fabricated memories, outdated memories, unresolved conflicts, or incorrect retrievals. The latter refers to hallucinations that arise during the generation phase of language models, where the model produces outputs inconsistent with factual truth or the given context. These two types of hallucinations are closely interrelated: memory hallucinations often act as upstream causes of generation hallucinations, while generation hallucinations may further amplify or obscure memory-related errors.

Generation hallucinations are the most extensively studied type of hallucination in current research and are typically divided into two categories: factual hallucinations and faithfulness hallucinations~\cite{huang2025survey}. Factual hallucinations assess whether the model output aligns with objective facts, whereas faithfulness hallucinations evaluate whether the output remains faithful to the given context or source information. Around these two categories, researchers have proposed a variety of mature evaluation methods and metric systems, including: external-retrieval-based factual verification~~\cite{DBLP:journals/corr/abs-2502-13622}, which compares generated content with external knowledge bases; model-internal-state-based reliability assessment~\cite{su-etal-2024-unsupervised,DBLP:conf/iclr/0026L0GWTFY24}, which analyzes attention distributions or activation patterns to estimate hallucination risk; behavior-based evaluation of output consistency and verifiability~\cite{liang-etal-2024-learning}; and uncertainty- or LLM-discriminator-based automated hallucination detection~\cite{kang2025uncertaintyquantificationhallucinationdetection,DBLP:journals/corr/abs-2505-08200}. These approaches have substantially improved the interpretability and quantifiability of hallucinations at the generation level, leading to relatively mature and systematic detection frameworks for generation hallucinations.

In contrast, research on hallucinations in memory systems is still in its infancy. Existing benchmarks such as LoCoMo, LongMemEval, PrefEval, and PersonaMem (Table~\ref{tab:memory-benchmarks-transposed}) focus on overall memory system performance rather than hallucination-specific evaluation.
The early benchmark LoCoMo~\cite{maharana2024evaluating} focuses on memory retention under long-context settings. It adopts an end-to-end evaluation paradigm, assessing models through question answering, summarization, and generation tasks to test factual recall and event tracking over ultra-long texts. Although the dataset is relatively large (about 7.5k questions), it only covers a time span of several months and lacks an explicit memory updating mechanism, primarily reflecting the model’s capability for static information retention.
Subsequently, LongMemEval~\cite{wulongmemeval} extends this framework by introducing metrics such as Information Retention Rate and Memory Recall Accuracy, covering approximately 2.5 years of multi-turn dialogue and incorporating explicit memory updates to quantify knowledge consistency across time. This represents a shift from static evaluation toward dynamic memory modeling.
In the direction of personalization, PrefEval~\cite{zhao2025do} evaluates a model’s ability to maintain and follow user preferences over long-term interactions, using generation and classification tasks to assess preference consistency, though it remains limited to persona-level memory. PersonaMem~\cite{jiang2025know} further constructs simulated user personas and event histories, employing multiple-choice evaluations to assess persona consistency, traceability, and update accuracy. With a longer time span (on the order of years), it provides a more representative benchmark for personalized long-term memory assessment.

Existing benchmarks primarily adopt holistic, end-to-end evaluations that treat memory systems as black boxes, making hallucinations only indirectly observable through final task performance rather than explicitly attributable to specific memory operations. In contrast, HaluMem is the first benchmark specifically designed to evaluate hallucinations in memory systems, enabling fine-grained, operation-level analysis and filling a critical gap left by prior memory evaluation benchmarks.

\section{Problem Definition}

\begin{figure*}[htp]
    \centering
    \includegraphics[width=0.8\linewidth]{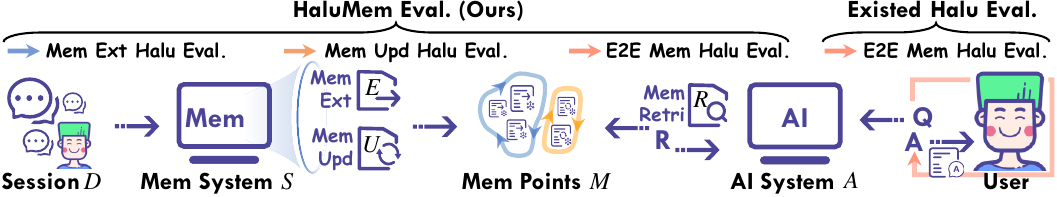}
    \caption{Comparison between HaluMem and existing benchmarks for memory systems.} 
    \label{fig:HaluMem_vs_ExistHaluMem}
\end{figure*}

Let there be a memory system $S$ that endows an AI system $A$ (including an LLM or AI agent) with long-term memory and personalization capabilities. The memory system receives a multi-turn dialogue sequence between the user and the assistant, denoted as $D = {(u_1,a_1),(u_2,a_2),\dots,(u_N,a_N)}$, where $u_i$ and $a_i$ represent the utterances of the user and the AI at turn $i$, respectively. Each memory point is stored as a plaintext entry, and a single memory is defined as $m$. With respect to the dialogue flow $D$, the memory system involves four types of operations during interaction:
(1) \textit{Memory Extraction} ($E$): extracting newly generated memory points from $D$;
(2) \textit{Memory Updating} ($U$): modifying or deleting existing memories;
(3) \textit{Memory Retrieval} ($R$): recalling memories relevant to the current query\footnote{Since retrieval $R$ primarily focuses on relevance and recall rate and rarely introduces generative processing by LLMs, this study concentrates on the three stages that directly induce hallucinations, namely $E$, $U$, and $Q$.};
(4) \textit{Memory Question Answering} ($Q$): constructing prompts and invoking $A$ to generate responses.

Existing evaluations of memory systems typically adopt an end-to-end question–answer paradigm. Given a set of dialogue-based queries $\mathcal{Q}=\{q_j\}_{j=1}^J$ and their corresponding gold answers $\mathcal{Y}^*=\{y_j^*\}_{j=1}^J$, the evaluation pipeline can be abstracted as
\[
\hat{M}=U\left(E(D)\right),\quad
\hat{R}_j=R(\hat{M},q_j),\quad
\hat{y}_j=A\left(\hat{R}_j,q_j\right).
\]
End-to-end evaluation is measured using answer-level metrics such as accuracy or F1 score:
\[
\text{Acc}_{\text{e2e}}=\frac{1}{J}\sum_{j=1}^J \mathbb{I}\left[\hat{y}_j=y_j^*\right].
\]

When $\hat{y}_j\neq y_j^*$, the metric $\text{Acc}_{\text{e2e}}$ cannot identify the source of the error. It remains unclear whether the hallucination arises from the extraction stage $E$, where incorrect or fabricated memories are introduced, from the updating stage $U$, where old memories are mistakenly modified or not properly refreshed, or from the question-answering stage $Q$, where unsupported generative content is produced despite correct memories being available. The lack of traceability prevents the development of targeted mitigation strategies.

To enable a localized and diagnostic evaluation, we construct fine-grained annotations and define gold standards for each stage.
(1) \textit{Extraction gold standard:} $G^{\mathrm{ext}} = \{m_i\}_{i=1}^{K}$, representing the set of memory points that should be newly added during the dialogue.
(2) \textit{Updating gold standard:} $G^{\text{upd}}=\{m^{\text{old}}\to m^{\text{new}}\}$, representing the set of memory point pairs before and after updates during the dialogue.
(3) \textit{Question–answer dataset:} for each query $q_j$, a gold answer $y_j^*$ is provided.
The system outputs are defined as follows:
\[
\hat{M}^{\text{ext}}=E(D),\quad
\hat{G}^{\text{upd}}=U(\hat{M}^{\text{ext}},D),\quad
\hat{y}_j=A\left(R(\hat{M},q_j), q_j\right),
\]
where $\hat{M}$ denotes the set of memory points representing the current state of the memory system when query $q_j$ is processed. By providing stage-specific gold standards and evaluation metrics for $E$, $U$, and $Q$, the proposed \textbf{HaluMem} benchmark enables operation-level hallucination evaluation within memory systems.

\section{Methodology for Constructing HaluMem}

\begin{figure*}[htp]
    \centering
    \includegraphics[width=1.\linewidth]{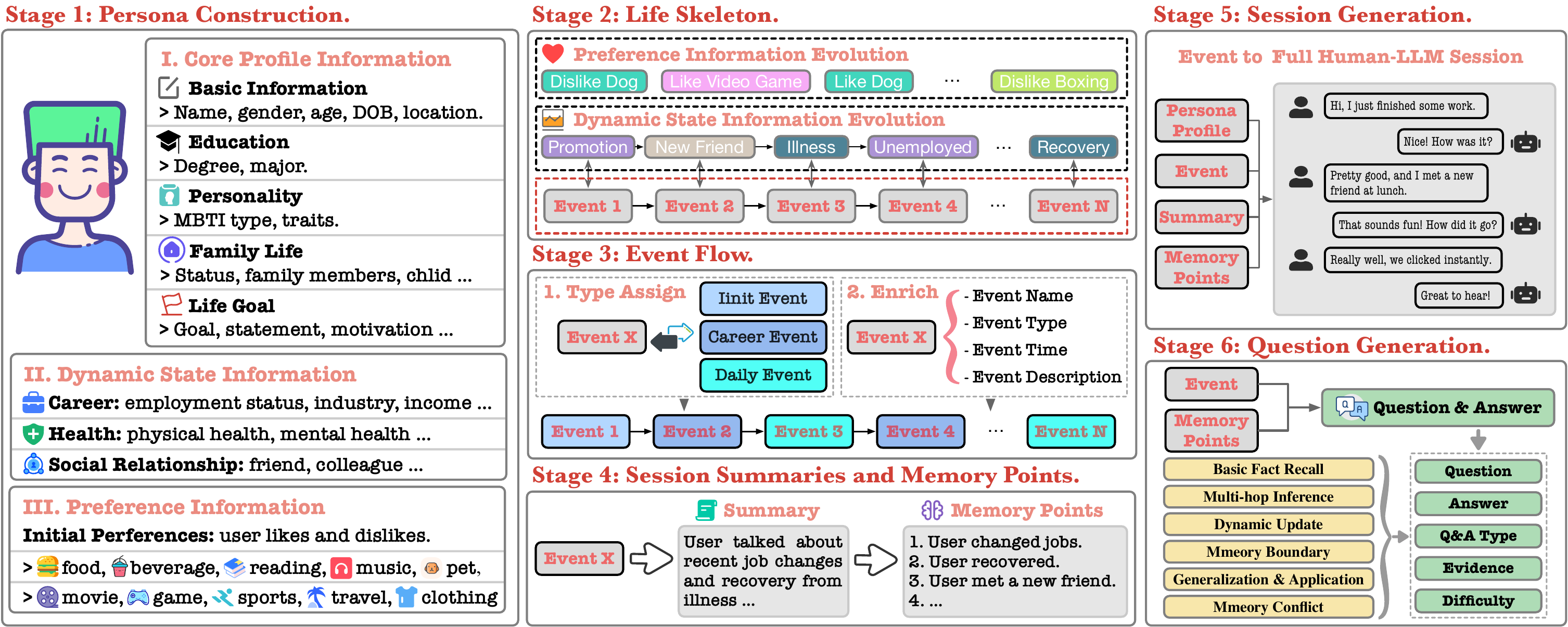}
    \caption{Framework of the HaluMem Construction Pipeline.} 
    \label{fig:framework}
\end{figure*}

To systematically evaluate memory systems in realistic scenarios, we construct the Hallucination in Memory Benchmark (HaluMem). To ensure the quality of the dataset and the controllability of the construction process, we design a user-centered, six-stage procedure based on a progressive expansion strategy.

\textbf{Stage 1: Persona Construction.}
This stage initiates the HaluMem dataset construction by creating virtual users with complete persona profiles to simulate real participants in later human–AI dialogues. Each persona includes three parts: \textit{Core Profile Information}, \textit{Dynamic State Information}, and \textit{Preference Information}. The core profile captures stable background traits; the dynamic state reflects current circumstances such as career, health, and relationships; and the preferences define personal tastes. Each user receives six to eight randomly assigned preferences across areas like food, music, and film. The core profile provides a static foundation, while the dynamic and preference elements, which evolve randomly, add realism, diversity, and rich material for memory extraction. An initial timestamp ensures that all personas reflect a consistent starting point in time. To enhance the authenticity of virtual users, user seeds are randomly sampled from the \textit{Persona Hub}\footnote{A collection of one billion diverse personas automatically curated from web data}~\cite{tao2024personahub}, and rule‑based procedures are applied to generate structured persona drafts. GPT‑4o then verifies and refines them. See Appendix~\ref{appendix:example of user profile} for an example.

\textbf{Stage 2: Life Skeleton.} After generating persona profiles, the second stage builds a \textit{life skeleton} that defines each user's evolutionary trajectory. Each user receives several core career events centered on life goals, which serve as anchors for the evolution of dynamic information. Updates to social status, career transitions, or health conditions are typically associated with these career events. Preference information evolves separately through probabilistic modifications or deletions, independent of these career events. These probabilistic rules ensure a diverse yet coherent evolution. The life skeleton captures the user's potential future states and serves as a structured script for later memory addition, modification, and deletion, maintaining the complexity and consistency of the evaluation scenarios.

\textbf{Stage 3: Event Flow.}
As the core component of dataset construction, the third stage aims to transform the abstract “life skeleton” generated in Stage 2 into a structured and narrative \textit{event flow}. The objective is to “eventify” discrete evolution instructions, constructing for each persona a complete memory timeline that integrates initial states, career development, and daily preference changes, thereby balancing narrative coherence with machine interpretability. The core of this stage includes three types of events:

\begin{itemize}
\item \textbf{Init Events:} Generated from the user’s initial profile, covering core, dynamic, and preference information. They serve as the starting point of the memory timeline, simulating the user’s first self-introduction.
\item \textbf{Career Events:} Derived from the life skeleton built in Stage~2, representing the main storyline of user development. Each career event is divided into sub-stages and instantiated with dynamic details (e.g., promotions, illnesses) to enrich the narrative.
\item \textbf{Daily Events:} Generated from the evolution of user preferences, independent of career progression. Each preference change becomes a concrete life scenario recorded as an atomic event with pre- and post-change states and their cause.
\end{itemize}

Within this framework, career events serve as the narrative backbone, while init and daily events provide necessary background and contextual details. Through the integration and chronological alignment of the three event types, this stage produces a coherent and complete event sequence that functions as the user’s \textit{memory transaction log}. See Appendix~\ref{appendix:example of event} for event examples.

\textbf{Stage 4: Session Summaries and Memory Points.} This stage transforms the structured event flow from Stage 3 into realistic session summaries and detailed memory points. For each event, we create a human–AI dialogue scenario shaped by the user’s motivation. The system has access to the current persona profile, along with all prior events and memory points, ensuring logical, causal, and consistent generation. As events unfold, the persona profile is dynamically updated to reflect the user's evolving state. Each memory point includes its content, type(persona, event, or relationship), and importance, with updated entries preserving replaced information for traceability. More details provided in Appendices~\ref{appendix:memory types} and~\ref{appendix:example of stage 4-6}.

\textbf{Stage 5: Session Generation.} This stage converts the structured event flow and memory points from previous stages into complete, multi-turn dialogues that are context-rich, goal-driven, and adversarially challenging. The process has three steps: adversarial content injection, multi-turn dialogue generation, and memory self-verification. Adversarial content injection adds distractor memories\footnote{False but similar memories that the AI naturally uses while the user stays silent, mimicking realistic information contamination}. Memory self-verification checks and refines each memory point for consistency with the generated dialogues. Overall, this stage simulates how memory is formed, maintained, and challenged in realistic conversations, producing data that test long-term memory performance and hallucination resistance. Examples appear in Appendix~\ref{appendix:example of stage 4-6}.

\textbf{Stage 6: Question Generation.}
The final stage constructs a set of memory-related question–answer pairs based on the sessions and memory points generated previously. Six categories of memory evaluation questions are predefined, and the number and types of questions are programmatically allocated according to event type and complexity to ensure balanced coverage. For each career event, all its sub-stages are integrated into a single unit to increase reasoning depth and complexity. Each question–answer pair is annotated with a difficulty level and accompanied by traceable evidence, explicitly linking the answer to the supporting memory points. See Appendices~\ref{appendix:question types} and ~\ref{appendix:example of stage 4-6} for details.

\textbf{Human Annotation.}
To verify the quality of \textit{HaluMem}, we conducted human annotation on part of the sessions in \textit{HaluMem-Medium}, covering both memory points and question–answer pairs. For each user, 35 sessions were randomly selected, totaling 700 sessions (over 50\% of the dataset). Eight annotators with at least a bachelor’s degree rated each session on \textit{Correctness}, \textit{Relevance}, and \textit{Consistency}. After 10 days of annotation, the results showed a correctness rate of 95.70\%, an average relevance score of 9.58, and an average consistency score of 9.45. These results demonstrate the high quality and reliability of the \textit{HaluMem} benchmark. More details are provided in Appendix~\ref{appendix: annotation}.

Overall, We constructed two datasets: \textit{HaluMem-Medium} and \textit{HaluMem-Long}.
\textit{HaluMem-Medium} includes 30,073 rounds of dialogue from 20 users, with an average context length of about 160k tokens, 14,948 memory points, and 3,467 QA pairs.
\textit{HaluMem-Long} extends each user’s context to 1M tokens through inserted irrelevant dialogues\footnote{Mainly sourced from ELI5~\cite{fan2019eli5},  \href{https://huggingface.co/datasets/Jackrong/GPT-OSS-120B-Distilled-Reasoning-math}{GPT-OSS-120B-Distilled-Reasoning-math}, and factual QA pairs generated using GPT-4o.}, containing 53,516 rounds in total. Details are given in Appendices~\ref{appendix:datset statistics} and~\ref{appendix:details of long}.

\section{Evaluation Framework of HaluMem}
\label{section: eval of halumem}

For each user, the session-level evaluation procedure of \textit{HaluMem} is defined as follows:
(1) Dialogue sessions $D^{1}, D^{2}, \dots, D^{S}$ are sequentially fed, in chronological order, into the memory system $S$.
(2) If the current session $D^{s}$ contains reference memory points or QA tasks, the corresponding evaluation process (extraction, updating, or question answering) is triggered immediately after $S$ completes processing that session, and the results are recorded.
(3) After processing all sessions, the metrics of the three categories of tasks are aggregated to obtain the overall system performance.

\begin{figure*}[htp]
    \centering
    \includegraphics[width=\linewidth]{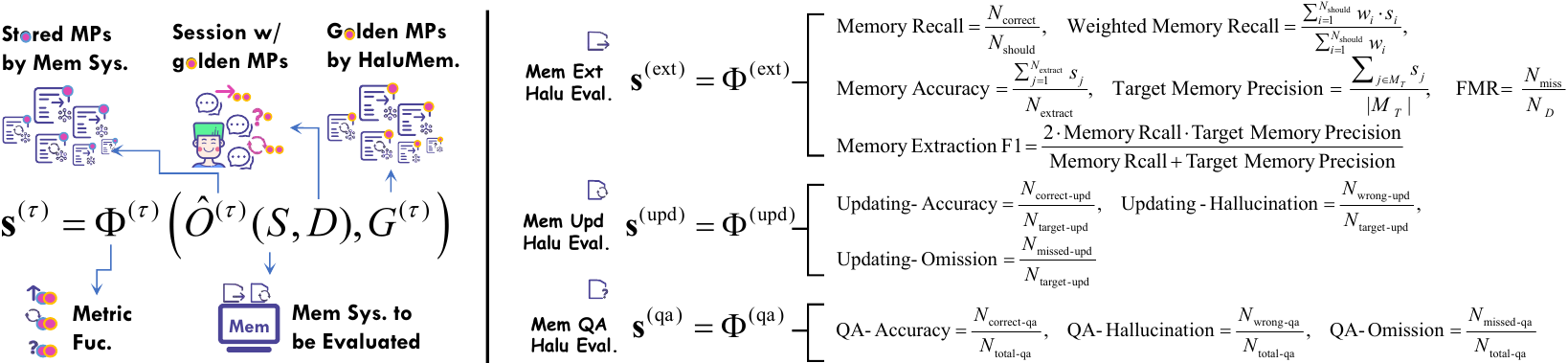}
    \caption{Hallucination evaluation process.}
    \label{fig:HaluMem_EvalF}
\end{figure*}

To support this evaluation workflow, the system is required to provide the following three types of API interfaces:
(1) \textbf{Add Dialogue API}: inputs dialogues and automatically performs memory extraction;
(2) \textbf{Get Dialogue Memory API}: retrieves the memory points extracted by the system from a specified session;
(3) \textbf{Retrieve Memory API}: retrieves the user's most relevant memory content based on a given query.

Based on the above workflow and interface design, \textit{HaluMem} conducts operation-level evaluation of memory systems across three essential tasks: \textit{Memory Extraction}, \textit{Memory Updating}, and \textit{Memory Question Answering}.

\subsection{Memory Extraction}

The memory extraction task evaluates whether the system can correctly identify and store key information from dialogues while avoiding fabricated or irrelevant memories.  
For each dialogue session $D^{s}$ that contains reference memories, the benchmark provides a gold memory set $G^{\mathrm{ext}}_{s} = \{m_i^{s}\}_{i=1}^{K_s}$ that should be extracted. The system output after processing $D^{s}$ is the extracted memory set $\widehat{M}^{\mathrm{ext}}_{s} = \{\hat{m}_j^{s}\}_{j=1}^{\widehat{K}_s}$, which is used for evaluation.

\paragraph{Memory Integrity (Anti-Amnesia)}  
This metric measures whether the system omits crucial information that should be extracted:
\begin{equation}
  \text{Memory Recall} = \frac{N_{\mathrm{correct}}}{N_{\mathrm{should}}}, \quad
  \text{Weighted Memory Recall} = \frac{\sum_{i=1}^{N_{\mathrm{should}}} w_i \cdot s_i}{\sum_{i=1}^{N_{\mathrm{should}}} w_i},
  \label{eq:mem-Integrity}
\end{equation}
where $N_{\mathrm{should}} = |G^{\mathrm{ext}}_{s}|$, $N_{\mathrm{correct}}$ denotes the number of correctly extracted memories, $w_i$ represents the importance weight of the $i$-th memory, and $s_i \in \{1, 0.5, 0\}$ indicates the extraction score (completed extracted, partially extracted, or omitted).

\paragraph{Memory Accuracy (Anti-Hallucination)}  
This metric evaluates whether the extracted memories are factual and free from hallucination:
\begin{equation}
  \text{Memory Accuracy} = \frac{\sum_{j=1}^{N_{\mathrm{extract}}} s_j}{N_{\mathrm{extract}}}, \quad
  \text{Target Memory Precision} = \frac{\sum_{j \in M_T} s_j}{|M_T|},
  \label{eq:mem-accuracy}
\end{equation}
where $N_{\mathrm{extract}} = |\widehat{M}^{\mathrm{ext}}_{s}|$, and $M_T \subset \widehat{M}^{\mathrm{ext}}_{s}$ denotes the set of target memories that match the reference ones.

\paragraph{False Memory Resistance (FMR)}  
This metric measures the system's ability to resist hallucination when facing distracting content that the AI mentions but the user does not confirm:
\begin{equation}
  \text{FMR} = \frac{N_{\mathrm{miss}}}{N_D},
  \label{eq:fmr}
\end{equation}
where $N_D$ represents the total number of distractor memories and $N_{\mathrm{miss}}$ denotes the number of distractors successfully ignored by the system, where a higher value indicates stronger resistance.

\paragraph{Memory Extraction F1}
We additionally report an F1 score to measure the overall performance of the memory extraction task by jointly considering completeness and correctness. Memory Recall ($R_{\text{mem}}$) is used as the recall term, while Target Memory Precision ($P_{\text{tgt}}$) is used as the precision term. The F1 score is defined as:

\begin{equation}
\mathrm{F1}_{\text{mem}} =
\frac{2\, R_{\text{mem}}\, P_{\text{tgt}}}
{R_{\text{mem}} + P_{\text{tgt}}}
\label{eq:mem-f1}
\end{equation}

\subsection{Memory Updating}

The memory updating task evaluates whether the system can correctly modify, merge, or replace existing memories during new dialogues so that consistency is maintained without introducing hallucinations.  
For each dialogue session $D^{s}$ that contains annotated updates, the gold update set is defined as $G^{\mathrm{upd}}_{s} = \{(m^{\mathrm{old}} \rightarrow m^{\mathrm{new}})\}$. The system output is denoted as \( \widehat{G}^{\mathrm{upd}}_{s} \).

Typical memory update hallucinations include: (1) incorrect modification of old information, (2) omission of new information, and (3) version conflicts or self-contradictions. Therefore, the following metrics are defined to evaluate memory update hallucination:
\begin{equation}
\begin{aligned}
  \text{Memory Updating Accuracy} &= \frac{N_{\mathrm{correct\text{-}upd}}}{N_{\mathrm{target\text{-}upd}}}, \\
  \text{Memory Updating Hallucination Rate} &= \frac{N_{\mathrm{wrong\text{-}upd}}}{N_{\mathrm{target\text{-}upd}}}, \\
  \text{Memory Updating Omission Rate} &= \frac{N_{\mathrm{missed\text{-}upd}}}{N_{\mathrm{target\text{-}upd}}},
\end{aligned}
\label{eq:mem-updating}
\end{equation}
where $N_{\mathrm{target\text{-}upd}} = |G^{\mathrm{upd}}_{s}|$, $N_{\mathrm{correct\text{-}upd}}$ is the number of correctly updated items, $N_{\mathrm{wrong\text{-}upd}}$ is the number of incorrect or hallucinated updates, and $N_{\mathrm{missed\text{-}upd}}$ is the number of updates that should have been made but were not.

\subsection{Memory Question Answering}

The memory question-answering task evaluates the end-to-end performance of the system, including extraction, updating, retrieval, and generation.  
For each question $q_j$, the system uses the \texttt{Retrieve Memory API} to obtain relevant memories $\widehat{R}(q_j)$. The retrieved set $\widehat{R}(q_j)$ and the question are then passed to the AI system $A$ to generate an answer $\hat{y}_j$. The generated answer is compared with the reference answer $y_j^*$, and the following metrics are defined:
\begin{equation}
\begin{aligned}
  \text{Memory QA Accuracy} &= \frac{N_{\mathrm{correct\text{-}qa}}}{N_{\mathrm{total\text{-}qa}}}, \\
  \text{Memory QA Hallucination Rate} &= \frac{N_{\mathrm{wrong\text{-}qa}}}{N_{\mathrm{total\text{-}qa}}}, \\
  \text{Memory QA Omission Rate} &= \frac{N_{\mathrm{missed\text{-}qa}}}{N_{\mathrm{total\text{-}qa}}},
\end{aligned}
\label{eq:mem-qa}
\end{equation}
where $N_{\mathrm{total\text{-}qa}}$ denotes the total number of questions, $N_{\mathrm{correct\text{-}qa}}$ denotes the number of correctly answered questions, $N_{\mathrm{wrong\text{-}qa}}$ denotes the number of questions answered with fabricated or incorrect information, and $N_{\mathrm{missed\text{-}qa}}$ refers to the number of questions that are left unanswered due to missing memories.

\section{Experiments}

\subsection{Experimental Setup}

We conducted a comprehensive evaluation of several state-of-the-art memory systems on HaluMem, including Mem0 (both standard and graph versions)~\cite{chhikara2025mem0}, Memobase~\cite{memobase}, MemOS~\cite{li2025memos_long}, Supermemory~\cite{supermemory}, and Zep~\cite{rasmussen2025zep}. Each memory system was independently evaluated in two subsets, HaluMem-Medium and HaluMem-Long, with efforts made to ensure consistent parameter configurations across evaluations.

To automate the evaluation of three core tasks, memory extraction, memory updating, and memory question answering, we use GPT-4o for consistency determination and scoring. We designed various prompt templates to guide the automated evaluation by GPT-4o (See Appendix~\ref{appendix:prompt for evaluation} for details.). In the memory updating task, we retrieved the 10 most relevant memories from the memory system for each memory labeled as "update type" for verification. In the memory question answering task, we retrieved 20 most relevant memories for each question to assist in generating answers, using GPT-4o uniformly as the answer generation model. The prompt templates used for answer generation across different memory systems are provided in Appendix~\ref{appendix: prompts for qa task}.

Some memory systems required specific Configurations due to their unique interfaces and constraints; details are provided in Appendix~\ref{appendix:special_settings}.

\subsection{Experimental Results}

Following the evaluation procedure outlined in Section~\ref{section: eval of halumem}, we conducted comprehensive evaluations of all memory systems across the three tasks in the HaluMem benchmark. The results were aggregated, and all metrics introduced in Section~\ref{section: eval of halumem} were subsequently computed.

\subsubsection{Overall Evaluation on HaluMem}

\begin{table}[!htp]
    \caption{Evaluation results of all memory systems on HaluMem. “R” denotes Recall, “Target P” denotes Target Memory Precision, “Acc.” denotes Accuracy, “FMR” denotes False Memory Resistance, “F1” denotes Memory Extraction F1-score, “C” denotes Correct Rate (Accuracy), “H” denotes Hallucination Rate, and “O” denotes Omission Rate. The values in parentheses in the “Target P” and “Acc.” columns represent the number of extracted memories. Color scale reflects performance (\colorbox{softred}{red} = worse, \colorbox{softgreen}{green} = better); Best values in \textbf{bold}.}
    \label{tab:evaluation results}
    \centering
    \resizebox{\linewidth}{!}{
    \begin{threeparttable}
        \begin{tabular}{cccccccccccccc}
        \toprule
        \multirow{3}{*}{\textbf{Dataset}} & 
        \multirow{3}{*}{\textbf{System}} & 
        \multicolumn{6}{c}{\textbf{Memory Extraction}} & 
        \multicolumn{3}{c}{\textbf{Memory Updating}} & 
        \multicolumn{3}{c}{\textbf{Question Answering}} \\
        \cmidrule(lr){3-8} \cmidrule(lr){9-11} \cmidrule(lr){12-14}
        & & \textbf{R$\uparrow$} & \textbf{Weighted R$\uparrow$} & \textbf{Target P$\uparrow$} & \textbf{Acc.$\uparrow$} & \textbf{FMR$\uparrow$} & \textbf{F1$\uparrow$} & \textbf{C$\uparrow$} & \textbf{H$\downarrow$} & \textbf{O$\downarrow$} & \textbf{C$\uparrow$} & \textbf{H$\downarrow$} & \textbf{O$\downarrow$}  \\
        \midrule
        \multirow{4}{*}{Medium} & Mem0 & \cellcolor{softbeige!91!softred}42.91\% & \cellcolor{softgreen!54!softbeige}65.03\% & \cellcolor{softgreen!84!softbeige}86.26\%(10556) & \cellcolor{softgreen!44!softbeige}60.86\%(16291) & \cellcolor{softgreen!34!softbeige}56.80\% & \cellcolor{softgreen!37!softbeige}57.31\% & \cellcolor{softbeige!70!softred}25.50\% & \cellcolor{softbeige!0!softgreen}0.45\% & \cellcolor{softred!69!softbeige}74.02\% & \cellcolor{softgreen!24!softbeige}53.02\% & \cellcolor{softbeige!61!softgreen}19.17\% & \cellcolor{softbeige!73!softgreen}27.81\% \\
         & Mem0-Graph & \cellcolor{softbeige!92!softred}43.28\% & \cellcolor{softgreen!54!softbeige}65.52\% & \cellcolor{softgreen!86!softbeige}87.20\%(10567) & \cellcolor{softgreen!46!softbeige}\textbf{61.86\%(16230)} & \cellcolor{softgreen!31!softbeige}55.70\% & \cellcolor{softgreen!37!softbeige}57.85\% & \cellcolor{softbeige!69!softred}24.50\% & \cellcolor{softbeige!0!softgreen}\textbf{0.26\%} & \cellcolor{softred!70!softbeige}75.24\% & \cellcolor{softgreen!28!softbeige}54.66\% & \cellcolor{softbeige!61!softgreen}19.28\% & \cellcolor{softbeige!72!softgreen}26.06\% \\
         & Memobase & \cellcolor{softbeige!52!softred}14.55\% & \cellcolor{softbeige!70!softred}25.88\% & \cellcolor{softgreen!91!softbeige}\textbf{92.24\%(5443)} & \cellcolor{softbeige!80!softred}32.29\%(17081) & \cellcolor{softgreen!77!softbeige}\textbf{80.78\%} & \cellcolor{softbeige!70!softred}25.13\% & \cellcolor{softbeige!31!softred}5.20\% & \cellcolor{softbeige!0!softgreen}0.55\% & \cellcolor{softred!93!softbeige}94.25\% & \cellcolor{softbeige!83!softred}35.33\% & \cellcolor{softbeige!76!softgreen}29.97\% & \cellcolor{softbeige!82!softgreen}34.71\% \\
         & MemOS & \cellcolor{softgreen!69!softbeige}\textbf{74.07\%} & \cellcolor{softgreen!82!softbeige}\textbf{84.81\%} & \cellcolor{softgreen!84!softbeige}86.25\%(45190) & \cellcolor{softgreen!42!softbeige}59.55\%(71793) & \cellcolor{softbeige!93!softred}44.94\% & \cellcolor{softgreen!76!softbeige}\textbf{79.70\%} & \cellcolor{softgreen!48!softbeige}\textbf{62.11\%} & \cellcolor{softbeige!0!softgreen}0.42\% & \cellcolor{softbeige!86!softgreen}\textbf{37.48\%} & \cellcolor{softgreen!58!softbeige}\textbf{67.23\%} & \cellcolor{softbeige!54!softgreen}\textbf{15.17\%} & \cellcolor{softbeige!58!softgreen}\textbf{17.59\%} \\
         & Supermemory & \cellcolor{softbeige!90!softred}41.53\% & \cellcolor{softgreen!52!softbeige}64.76\% & \cellcolor{softgreen!89!softbeige}90.32\%(14134) & \cellcolor{softgreen!44!softbeige}60.83\%(22551) & \cellcolor{softgreen!14!softbeige}51.77\% & \cellcolor{softgreen!34!softbeige}56.90\% & \cellcolor{softbeige!56!softred}16.37\% & \cellcolor{softbeige!14!softgreen}1.15\% & \cellcolor{softred!80!softbeige}82.47\% & \cellcolor{softgreen!28!softbeige}54.07\% & \cellcolor{softbeige!66!softgreen}22.24\% & \cellcolor{softbeige!67!softgreen}23.69\% \\
         & Zep & - & - & - & - & - & - & \cellcolor{softbeige!96!softred}47.28\% & \cellcolor{softbeige!0!softgreen}0.42\% & \cellcolor{softred!20!softbeige}52.31\% & \cellcolor{softgreen!31!softbeige}55.47\% & \cellcolor{softbeige!64!softgreen}21.92\% & \cellcolor{softbeige!66!softgreen}22.62\% \\
        \midrule
        \multirow{4}{*}{Long} & Mem0 & \cellcolor{softbeige!24!softred}3.23\% & \cellcolor{softbeige!46!softred}11.89\% & \cellcolor{softgreen!87!softbeige}88.01\%(1134) & \cellcolor{softbeige!95!softred}\textbf{46.01\%(2433)} & \cellcolor{softgreen!86!softbeige}87.65\% & \cellcolor{softbeige!34!softred}6.22\% & \cellcolor{softbeige!14!softred}1.45\% & \cellcolor{softbeige!0!softgreen}\textbf{0.03\%} & \cellcolor{softred!97!softbeige}98.51\% & \cellcolor{softbeige!74!softred}28.11\% & \cellcolor{softbeige!58!softgreen}17.29\% & \cellcolor{softred!28!softbeige}54.60\% \\
         & Mem0-Graph & \cellcolor{softbeige!20!softred}2.24\% & \cellcolor{softbeige!44!softred}10.76\% & \cellcolor{softgreen!86!softbeige}87.32\%(785) & \cellcolor{softbeige!90!softred}41.26\%(1866) & \cellcolor{softgreen!87!softbeige}\textbf{88.36\%} & \cellcolor{softbeige!28!softred}4.36\% & \cellcolor{softbeige!14!softred}1.47\% & \cellcolor{softbeige!0!softgreen}0.04\% & \cellcolor{softred!97!softbeige}98.40\% & \cellcolor{softbeige!80!softred}32.44\% & \cellcolor{softbeige!64!softgreen}21.82\% & \cellcolor{softbeige!94!softgreen}45.74\% \\
         & Memobase & \cellcolor{softbeige!34!softred}6.18\% & \cellcolor{softbeige!52!softred}14.68\% & \cellcolor{softgreen!87!softbeige}\textbf{88.56\%(3077)} & \cellcolor{softbeige!70!softred}25.61\%(11795) & \cellcolor{softgreen!83!softbeige}85.39\% & \cellcolor{softbeige!46!softred}11.55\% & \cellcolor{softbeige!28!softred}4.10\% & \cellcolor{softbeige!0!softgreen}0.36\% & \cellcolor{softred!94!softbeige}95.38\% & \cellcolor{softbeige!81!softred}33.60\% & \cellcolor{softbeige!76!softgreen}29.46\% & \cellcolor{softbeige!84!softgreen}36.96\% \\
         & MemOS & \cellcolor{softgreen!78!softbeige}\textbf{81.90\%} & \cellcolor{softgreen!88!softbeige}\textbf{89.56\%} & \cellcolor{softgreen!80!softbeige}82.32\%(48246) & \cellcolor{softbeige!92!softred}43.77\%(99462) & \cellcolor{softbeige!74!softred}28.85\% & \cellcolor{softgreen!80!softbeige}\textbf{82.11\%} & \cellcolor{softgreen!54!softbeige}\textbf{65.25\%} & \cellcolor{softbeige!0!softgreen}0.29\% & \cellcolor{softbeige!82!softgreen}\textbf{34.47\%} & \cellcolor{softgreen!52!softbeige}\textbf{64.44\%} & \cellcolor{softbeige!56!softgreen}\textbf{16.61\%} & \cellcolor{softbeige!60!softgreen}\textbf{18.95\%} \\
         & Supermemory & \cellcolor{softgreen!24!softbeige}53.02\% & \cellcolor{softgreen!63!softbeige}70.73\% & \cellcolor{softgreen!83!softbeige}85.82\%(24483) & \cellcolor{softbeige!76!softred}29.71\%(77134) & \cellcolor{softbeige!84!softred}36.86\% & \cellcolor{softgreen!54!softbeige}65.54\% & \cellcolor{softbeige!58!softred}17.01\% & \cellcolor{softbeige!0!softgreen}0.58\% & \cellcolor{softred!80!softbeige}82.42\% & \cellcolor{softgreen!24!softbeige}53.77\% & \cellcolor{softbeige!66!softgreen}22.21\% & \cellcolor{softbeige!69!softgreen}24.02\% \\
         & Zep & - & - & - & - & - & - & \cellcolor{softbeige!86!softred}37.35\% & \cellcolor{softbeige!0!softgreen}0.48\% & \cellcolor{softred!48!softbeige}62.14\% & \cellcolor{softgreen!0!softbeige}50.19\% & \cellcolor{softbeige!66!softgreen}22.51\% & \cellcolor{softbeige!73!softgreen}27.30\% \\
        \bottomrule
        \end{tabular}
        \begin{tablenotes}
            \footnotesize
            \item Note: since Zep does not provide a Get Dialogue Memory API, metrics related to memory extraction cannot be computed. For details, see Appendix~\ref{appendix:special_settings}.
        \end{tablenotes}
    \end{threeparttable}
    }
\end{table}

Table~\ref{tab:evaluation results} presents the evaluation results of all memory systems on three tasks: memory extraction, memory updating, and memory question answering. The evaluation metrics for memory extraction include memory integrity and memory accuracy.

Overall, most memory systems perform worse on HaluMem-Long than on HaluMem-Medium, with Mem0, Mem0-Graph, and Memobase showing particularly notable declines. Notably, the Mem0 series and Memobase extract significantly fewer memories on HaluMem-Long than on HaluMem-Medium, whereas Supermemory and MemOS exhibit the opposite trend. This indicates that \textbf{future memory systems need to improve their ability to process irrelevant information and distinguish between high- and low-value memories.}

In the memory extraction task, regarding memory integrity, except for MemOS, all systems achieve recall (R) rates below 60\%, indicating that many reference memory points are not extracted. The higher weighted memory recall (Weighted R) suggests that these systems can prioritize important memory points. Regarding memory accuracy, all systems have accuracy (Acc.) below 62\%, reflecting a high proportion of hallucinations, although performance on target memory precision (Target~P) is relatively good. Supermemory and MemOS perform the worst on FMR because they tend to extract excessive information without effectively filtering distractions or unhelpful content. Other systems adopt more conservative strategies and thus perform better in FMR. In terms of the F1 score, MemOS and Supermemory perform the best and exhibit stability in long contexts, whereas other systems experience a sharp decline. In summary, \textbf{future memory systems should strike a balance among coverage of important memories, extraction accuracy, and resistance to interference, aiming for both high quality and reliability in memory retrieval.}

In the memory updating task, most systems perform poorly, and their performance drops considerably on HaluMem-Long. Systems showing better performance in memory integrity also tend to exhibit higher update accuracy, but most systems suffer omission rates above 50\%. This issue primarily stems from insufficient coverage in memory extraction: when the pre-update memories are not extracted, related updates cannot be properly processed. Moreover, the fact that all systems exhibit hallucination rates below 2\% does not necessarily imply strong hallucination suppression, since very few samples actually enter the update stage. Overall, \textbf{current systems face a clear bottleneck in memory updating: the extraction and updating stages lack stable linkage, resulting in low accuracy and high omission rates.}

\begin{table}[!ht]
    \caption{Typewise accuracy on event, persona, and relationship memory.}
    \label{tab:Typewise Accuracy}
    \centering
    \resizebox{0.6\linewidth}{!}{
    \begin{threeparttable}
        \begin{tabular}{ccccc}
        \toprule
        \textbf{Dataset} & \textbf{System} & \textbf{Event} & \textbf{Persona} & \textbf{Relationship} \\
        \midrule
        \multirow{4}{*}{Medium} & Mem0 & 29.69\% & 33.74\% & 27.77\% \\
         & Mem0-Graph & 30.02\% & 33.71\% & 26.60\% \\
         & Memobase & 5.12\% & 13.38\% & 6.79\% \\
         & MemOS & \textbf{63.41\%} & \textbf{59.77\%} & \textbf{62.40\%} \\
         & Supermemory & 28.66\% & 32.11\% & 20.67\% \\
         & Zep & \underline{44.83\%$^*$} & \underline{49.75\%$^*$} & \underline{38.81\%$^*$} \\
        \midrule
        \multirow{4}{*}{Long} & Mem0 & 0.92\% & 3.01\% & 2.18\% \\
         & Mem0-Graph & 1.10\% & 2.00\% & 1.59\% \\
         & Memobase & 4.09\% & 5.32\% & 4.21\% \\
         & MemOS & \textbf{70.92\%} & \textbf{68.35\%} & \textbf{71.68\%} \\
         & Supermemory & \underline{38.48\%} & \underline{40.85\%} & \underline{32.61\%} \\
         & Zep & 35.76\%$^*$ & 39.07\%$^*$ & 31.16\%$^*$ \\
         \bottomrule
        \end{tabular}
        \begin{tablenotes}
            \footnotesize
            \item *~The memory entries of Zep include only those from the memory updating task. For details, see Appendix~\ref{appendix:special_settings}.
        \end{tablenotes}
    \end{threeparttable}
    }
\end{table}

In the memory question-answering task, the best-performing systems are also those that perform well in memory integrity and memory updating, further highlighting the crucial role of memory extraction. For example, Mem0 and Mem0-Graph show clear performance declines on HaluMem-Long compared to HaluMem-Medium, which strongly correlates with their substantial reduction in extracted memory points. However, all systems achieve answer accuracies below 70\%, with both hallucination rate and omission rate remaining high, and their overall performance further decreases on HaluMem-Long. This demonstrates that \textbf{current memory systems' QA performance depends heavily on the sufficiency and accuracy of upstream memory extraction, and remains prone to factual deviation and memory confusion under interference or extended context conditions.}

\begin{figure*}[htp]
    \centering
    \includegraphics[width=.9\linewidth]{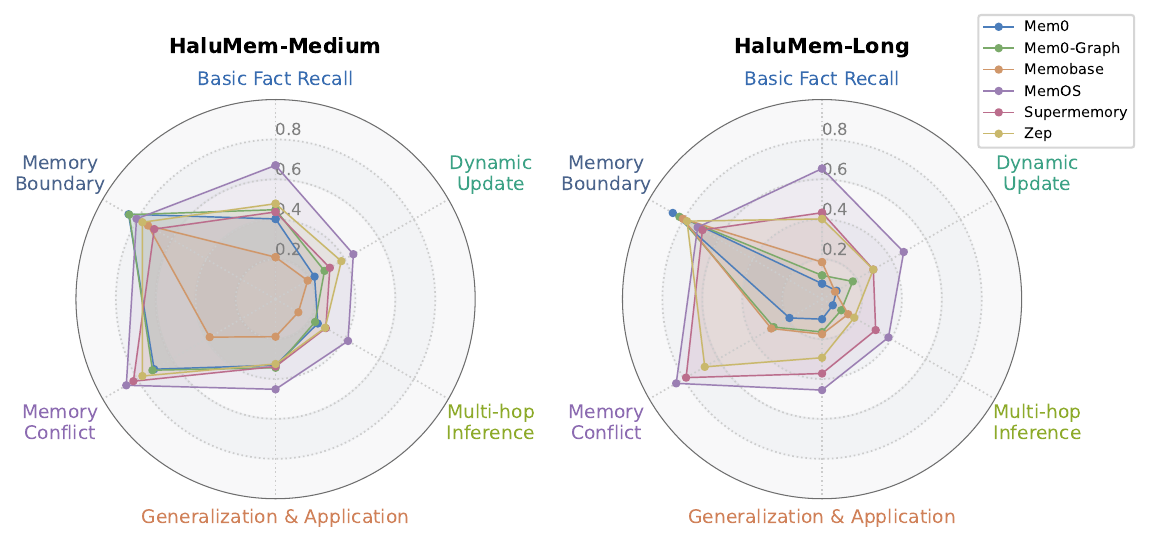}
    \caption{Performance of the Memory System Across Different Question Types} 
    \label{fig:radar}
\end{figure*}

\subsubsection{Performance on Different Memory Types}

Table~\ref{tab:Typewise Accuracy} reports the extraction accuracy of each memory system for event, persona, and relationship memories, which include all memory points from both the memory extraction and updating tasks while excluding distractor memories. MemOS achieves the best overall performance. However, Mem0, Mem0-Graph, and Memobase show a marked decline in long-context scenarios, suggesting difficulty in consistently capturing valuable information in complex dialogues. MemOS and Supermemory perform better on HaluMem-Long than on HaluMem-Medium, probably because they extract a larger number of memory points in the long-context condition. Across memory types, Persona memories yield slightly higher accuracy, indicating that static personal traits are easier to capture, whereas understanding event dynamics and relationship changes remains challenging. \textbf{Overall, all systems still show low performance across the three memory categories, indicating significant limitations in current memory modeling.}

\subsubsection{Performance on Different Question Types}

Figure~\ref{fig:radar} illustrates the performance of different memory systems across six categories of questions. \textbf{Overall, the accuracy of all memory systems remains relatively low across most categories, indicating substantial room for improvement.} The Mem0 series and Memobase show significantly poorer performance on HaluMem-Long compared to HaluMem-Medium, suggesting a notable degradation under ultra-long context conditions. In contrast, MemOS, SuperMemory, and Zep demonstrate relatively stable behavior and achieve consistently superior overall performance on both datasets. Furthermore, all memory systems perform comparatively better on memory boundary and memory conflict questions, \textbf{indicating their capability to effectively recognize unknown or misleading information and respond correctly.} However, their performance deteriorates substantially on multi-hop inference, dynamic update, and generalization \& application questions, \textbf{suggesting that current memory systems still struggle with complex reasoning and preference tracking.}

\subsubsection{Efficiency Analysis of Memory Systems}

\begin{table}[!htp]
    \caption{Time consumption of all memory systems during evaluation.}
    \centering
    \resizebox{0.7\linewidth}{!}{
    \begin{threeparttable}
        \begin{tabular}{ccccc}
        \toprule
        \multirow{2}{*}{\textbf{Dataset}} & \multirow{2}{*}{\textbf{System}} & \textbf{Dialogue Addition} & \textbf{Memory Retrieval} & \textbf{Total} \\
         & & \textbf{Time (min)} & \textbf{Time (min)}  & \textbf{Time (min)} \\
        \midrule
        \multirow{4}{*}{\textbf{Medium}} & Mem0 & 2768.14 & \underline{41.66} & 2809.8 \\
         & Mem0-Graph & 2840.07 & 54.65 & 2894.72 \\
         & Memobase & \underline{293.30} & 139.95 & \underline{433.25} \\
         & MemOS & 1028.84 & \textbf{20.52} & 1049.37 \\
         & Supermemory & \textbf{273.21} & 95.53 & \textbf{368.74} \\
         & Zep & - & 53.34 & - \\
        \midrule
        \multirow{4}{*}{\textbf{Long}} & Mem0 & \underline{691.62} & \underline{39.15} & \underline{730.77} \\
         & Mem0-Graph & 870.32 & 62.42 & 932.74 \\
         & Memobase & \textbf{239.29} & 136.19 & \textbf{375.48} \\
         & MemOS & 1524.39 & \textbf{20.96} & 1545.34 \\
         & Supermemory & 1672.53 & 137.02 & 1809.55 \\
         & Zep & - & 50.22 & - \\
        \bottomrule
        \end{tabular}
    \end{threeparttable}
    }
    \label{tab:Time consumption}
\end{table}	

Table~\ref{tab:Time consumption} shows the time consumption of all memory systems during the evaluation process for dialogue addition and memory retrieval, as well as their total runtime. Overall, dialogue addition requires substantially more time than memory retrieval, \textbf{indicating that the write stage is the primary computational bottleneck. Enhancing the efficiency of memory extraction and updating is thus crucial for improving interactive performance.} On HaluMem-Medium, Supermemory performs best in both dialogue addition and total runtime, while MemOS shows the best retrieval efficiency. However, the dialogue addition time of Mem0 and Mem0-Graph exceeds 2700 minutes, revealing their low processing efficiency during dialogue ingestion and memory construction. On HaluMem-Long, the dialogue addition time for Mem0, Mem0-Graph, and Memobase decreases, mainly because the number of processed memory points is reduced rather than due to performance improvement. In contrast, MemOS and Supermemory extract a substantially larger number of memory points, resulting in a significant increase in their time cost.

\section{Conclusion}

Most existing benchmarks for memory systems adopt a black box, end to end question answering setup, which makes it difficult to analyze and measure hallucinations introduced by internal memory operations. To address this gap, we present the Hallucination in Memory Benchmark (HaluMem), the first operation level hallucination evaluation benchmark for memory systems. HaluMem conducts a comprehensive assessment of memory hallucinations and overall performance through three tasks: memory extraction, memory updating, and memory question answering. For dataset construction, we design a user-centric, six-stage pipeline based on a progressive expansion strategy, and build two datasets, HaluMem-Medium and HaluMem-Long, whose construction quality is verified through human annotation. In the experimental study, we systematically evaluate multiple advanced memory systems on HaluMem, analyzing performance on the three tasks, extraction accuracy across different memory types, and efficiency. The results reveal persistent bottlenecks in coverage, accuracy, update capability, robustness to interference, and question answering reliability. Future work should improve extraction quality, update logic, semantic understanding, and system efficiency in order to achieve more stable and comprehensive long term memory.

\bibliographystyle{plainnat}
\bibliography{main}

\clearpage

\newpage
\appendix
\appendixpage

\startcontents[sections]
\printcontents[sections]{l}{1}{\setcounter{tocdepth}{2}}

\clearpage

\section{Supplementary Details of HaluMem}
\label{appendix:details of halumem}

This appendix provides additional statistical information and key definitions of the HaluMem dataset to support a more detailed understanding of its data composition and task taxonomy. The HaluMem dataset consists of two parts: \textbf{HaluMem-Medium} and \textbf{HaluMem-Long}, representing medium- and long-context multi-turn human–AI interaction scenarios, respectively. Each subset contains multiple types of memory points and questions, enabling systematic evaluation of hallucination behaviors in memory systems.

\begin{table}[!ht]
\centering
\caption{Statistical Overview of HaluMem Datasets}
\label{tab:halumem-stats}
\begin{tabular}{lcc}
\toprule
\textbf{Metrics} & \textbf{HaluMem-Medium} & \textbf{HaluMem-Long} \\
\midrule
\multicolumn{3}{l}{\textbf{Interaction Statistics}} \\
\lightmidrule
Avg Context Length (tokens/user) & 159,910.95 & 1,007,264.65 \\
Avg Session Num (per user) & 69.35 & 120.85 \\
Avg Dialogue Turns per Session & 21.68 & 22.14 \\
Total Dialogue Turns & 30,073 & 53,516 \\
\midrule
\multicolumn{3}{l}{\textbf{Memory Statistics}} \\
\lightmidrule
Avg Memory Num per Session & 10.78 & 6.18 \\
Distractor Memories & 2,648 & 2,648 \\
Update Memories & 3,122 & 3,122 \\
Persona Memories & 9,116 & 9,116 \\
Event Memories & 4,550 & 4,550 \\
Relationship Memories & 1,282 & 1,282 \\
Total Memories & 14,948 & 14,948 \\
\midrule
\multicolumn{3}{l}{\textbf{Question Statistics}} \\
\lightmidrule
Avg Questions per User & 173.35 & 173.35 \\
Total Questions & 3,467 & 3,467 \\
\textit{Question Type Distribution:} & & \\
\quad Basic Fact Recall & 746 & 746 \\
\quad Multi-hop Inference & 198 & 198 \\
\quad Dynamic Update & 180 & 180 \\
\quad Memory Boundary & 828 & 828 \\
\quad Memory Conflict & 769 & 769 \\
\quad Generalization \& Application & 746 & 746 \\
\bottomrule
\end{tabular}
\end{table}

\subsection{Definition of Memory Types}
\label{appendix:memory types}

HaluMem categorizes memory content into three core types, reflecting different semantic levels and stability characteristics:

\begin{itemize}
    \item \textbf{Persona Memory}: Describes user's identity, interests, habits, beliefs, and other stable characteristics.
    \item \textbf{Event Memory}: Records specific events, experiences, or plans that occurred to the user.
    \item \textbf{Relationship Memory}: Describes user's relationships, interactions, or views of others.
\end{itemize}

\subsection{Definition of Question Types}
\label{appendix:question types}

To comprehensively cover different types of hallucination, HaluMem defines six categories of evaluation questions:

\begin{itemize}
    \item \textbf{Basic Fact Recall}: Directly ask about single objective facts or user preferences that explicitly appear in the dialogue, without requiring reasoning or information integration.
    \item \textbf{Multi-hop Inference}: Requires synthesizing multiple information fragments from dialogues, and can only derive answers through logical reasoning or temporal reasoning.
    \item \textbf{Dynamic Update}: Tests the ability to track information changes over time, requiring identification of the latest status or preference changes.
    \item \textbf{Memory Boundary}: Tests the system's ability to identify unknown information by asking about details not mentioned in the input information to examine whether the system will fabricate answers.
    \item \textbf{Generalization \& Application}: Based on known user preferences or characteristics, infer reasonable suggestions or judgments in new scenarios.
    \item \textbf{Memory Conflict}: Tests the system's ability to identify and correct erroneous premises. Questions deliberately contain incorrect information that directly contradicts known memory points, requiring the system to identify contradictions, correct errors, and answer based on correct information.
\end{itemize}

\subsection{Dataset Statistics}
\label{appendix:datset statistics}

Table~\ref{tab:halumem-stats} presents the main statistical features of HaluMem-Medium and HaluMem-Long, including context scale, session quantity, memory distribution, and question-type composition. All values are based on the finalized dataset version.

\subsection{Construction Details of HaluMem‑Long}
\label{appendix:details of long}

HaluMem‑Long is built upon HaluMem‑Medium to test memory systems under ultra‑long context scenarios, focusing on robustness and hallucination suppression. Based on each user’s sessions in HaluMem‑Medium, additional irrelevant dialogues were inserted:

\begin{itemize}
    \item Within sessions: extra unrelated exchanges were added to existing conversations.
    \item Between sessions: new sessions composed entirely of irrelevant dialogues were interleaved.
\end{itemize}

These irrelevant dialogues include:

\begin{itemize}
    \item Factual Q\&A derived partly from the ELI5 dataset~\cite{fan2019eli5} and partly generated by us.
    \item Mathematical reasoning Q\&A adopted from \href{https://huggingface.co/datasets/Jackrong/GPT-OSS-120B-Distilled-Reasoning-math}{GPT-OSS-120B-Distilled-Reasoning-math}.
\end{itemize}

The ELI5 dataset consists of factual question–answer pairs (e.g., the second QA example), whereas GPT-OSS-120B-Distilled-Reasoning-math contains question–answer pairs involving mathematics (e.g., the third QA example). To further enrich the diversity of irrelevant dialogues, we also sampled factual QA pairs across eight domains using GPT-4o (e.g., the first example), including Historical Figure, Scientific Concept, Country or Place, Famous Invention, Philosophical Theory, Artwork or Painting, Historical Event, and Mathematical Theorem. These QA pairs are used to simulate dialogues between users and the AI driven by instrumental needs in realistic scenarios. They have minimal impact on the user’s original conversations and do not affect the memory system’s personalized memories of the user. See Appendix~\ref{appendix:example of irrelevant dialogue} for examples of irrelevant dialogues.

\section{Special Configurations for Some Memory Systems}
\label{appendix:special_settings}

This appendix documents the special configurations applied to several memory systems evaluated on HaluMem.
While the experimental setup strives to maintain consistent configurations across all evaluated systems, certain memory systems exhibit unique API constraints that necessitate specific adjustments or workarounds.
Each subsection below outlines these system-specific configurations to ensure reproducibility.

\subsection{Memobase}

Since Memobase does not provide a Get Dialogue Memory API, we adopted a localized deployment approach and directly accessed the corresponding dialogue memories from its underlying database. Additionally, the Retrieve Memory API of Memobase only supports controlling the maximum length of the returned memory text. Based on test results, we set the maximum length for memory recall in the memory updating task to 250 tokens and the recall length for the memory question answering task to 500 tokens.

\subsection{Zep}

According to our current understanding, the official APIs provided by Zep do not support retrieving all memory points within a specific session, meaning they do not offer functionality equivalent to a Get Dialogue Memory API. Consequently, we were unable to evaluate Zep’s performance on the memory extraction task. We attempted to use the function `thread.get\_user\_context()` offered by Zep to obtain all memories under a given thread; however, this method only returns recent memories rather than the complete set, which does not meet the evaluation requirements. Moreover, since Zep’s memory processing workflow operates entirely asynchronously, we could not accurately measure the time consumption in the dialogue addition phase and instead recorded only the time cost associated with memory retrieval.

\section{Annotation Guidelines and Instructions}
\label{appendix: annotation}

\subsection{Annotation Objective}
\textbf{Task Background:} Given a user's persona description and multi-turn human-AI dialogue content, memory points and question-answer (QA) pairs are generated using large language models. The generated items must be manually verified to ensure strict grounding in the dialogue content. Specifically, memory points should have explicit evidence in the dialogue, and QA pairs should be relevant to the dialogue, with answers directly inferable from it.  

\textbf{Core Objective:} Assess whether the content in the \textit{Evaluation Item} is consistent with the corresponding \textit{Dialogue Info}.  

\begin{figure*}[htp]
    \centering
    \includegraphics[width=1.\linewidth]{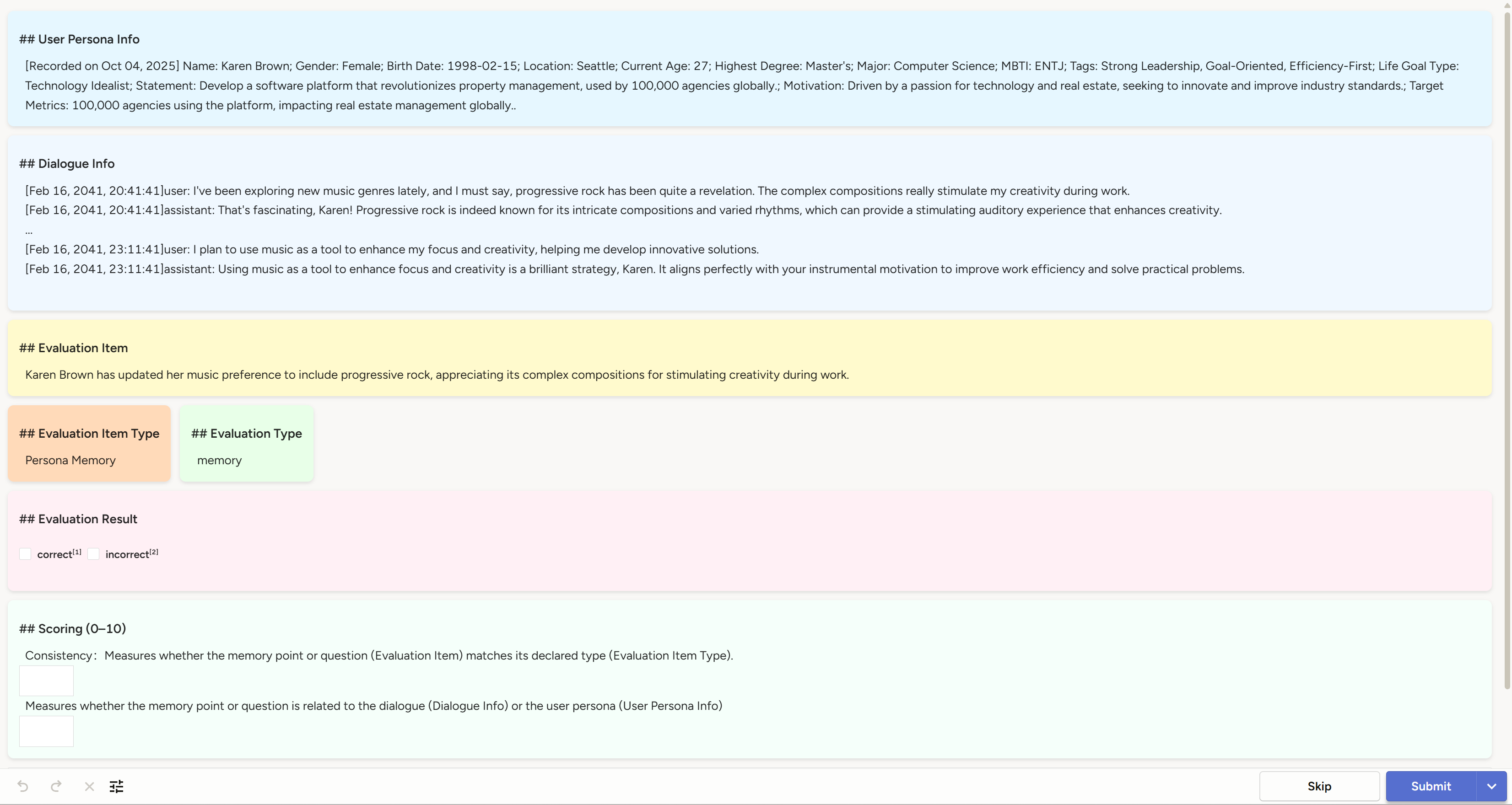}
    \caption{Annotation interface.} 
    \label{fig:annotation_interface}
\end{figure*}

An illustrative screenshot of the annotation interface is provided below (Figure~\ref{fig:annotation_interface}).

\subsection{Information Fields}
\begin{itemize}
    \item \textbf{User Persona Info}: Basic information about the user provided in the dialogue setting.
    \item \textbf{Dialogue Info}: Multi-turn dialogue content between the user and the AI. Each turn contains one user utterance (\textit{user}) and one assistant response (\textit{assistant}).
    \item \textbf{Evaluation Item}: The item to be annotated, which can be either a memory point or a QA pair, as indicated by the \textit{Evaluation Type}. For memory points, the item is a textual description about the user. For QA pairs, it includes a question and an answer (e.g., \textit{Question: xxx; Answer: xxx}).
    \item \textbf{Evaluation Type}: Indicates the type of \textit{Evaluation Item}: ``memory'' for memory points and ``question'' for QA pairs.
    \item \textbf{Evaluation Item Type}: Categorizes the memory point or question as follows:
    \begin{itemize}
        \item \textbf{Memory Points:}
        \begin{itemize}
            \item \textit{Persona Memory}: Describes user's identity, interests, habits, beliefs, and other stable characteristics.
            \item \textit{Event Memory}: Records specific events, experiences, or plans that occurred to the user.
            \item \textit{Relationship Memory}: Describes user's relationships, interactions, or perspectives on others.
        \end{itemize}
        \item \textbf{Questions:}
        \begin{itemize}
            \item \textit{Basic Fact Recall}: Directly asks about single objective facts or user preferences explicitly mentioned in the dialogue, without requiring reasoning or information integration.
            \item \textit{Multi-hop Inference}: Requires synthesizing multiple pieces of dialogue information, deriving answers through logical or temporal reasoning.
            \item \textit{Dynamic Update}: Tests the ability to track information changes over time, requiring identification of the latest status or preference changes.
            \item \textit{Memory Boundary}: Tests the system's ability to recognize unknown information by querying details not mentioned in the input, assessing whether the system will fabricate answers.
            \item \textit{Generalization \& Application}: Infers reasonable suggestions or judgments in new scenarios based on known user preferences or characteristics.
            \item \textit{Memory Conflict}: Evaluates the system's ability to identify and correct erroneous premises. Questions deliberately contain incorrect information contradicting known memory points, requiring the system to identify contradictions, correct errors, and answer based on correct information.
        \end{itemize}
    \end{itemize}
\end{itemize}

\subsection{Annotation Dimensions and Scoring}
Each memory point and QA pair is evaluated along three dimensions: \textit{Correctness}, \textit{Relevance}, and \textit{Consistency}.  

\begin{itemize}
    \item \textbf{Evaluation Result}: A single-choice judgment of ``correct'' or ``incorrect''. For memory points, this assesses whether the item is supported by the dialogue. For QA pairs, it assesses whether the question and answer can be clearly found in the dialogue.
    \item \textbf{Scoring (0--10)}: Two separate scores are assigned:
    \begin{itemize}
        \item \textit{Consistency}: Measures whether the memory point or question (\textit{Evaluation Item}) matches its declared type (\textit{Evaluation Item Type}). 0--3 indicates poor consistency, 4--6 partial consistency, and 7--10 full consistency.
        \item \textit{Relevance}: Measures whether the memory point or question is related to the dialogue (\textit{Dialogue Info}) or the user persona (\textit{User Persona Info}). 0--3 indicates low relevance, 4--6 moderate relevance, and 7--10 high relevance.
    \end{itemize}
\end{itemize}

\section{Prompts}
\label{appendix:prompts}

This section presents some of the important prompt templates involved in the paper.

\subsection{Prompts for Memory Question Answering Task}
\label{appendix: prompts for qa task}

Figures~\ref{img:Prompt for Mem0 and Mem0-Graph} $\sim$ \ref{img:Prompt for Zep} show the prompt templates used by all memory systems in memory question answering task to assemble questions and retrieve memory points, which are then fed into GPT-4o to generate responses. All of these memory templates are obtained from the official GitHub repositories of the respective memory systems.

\begin{figure}[!htp]
    \small
    \centering
    \begin{tcolorbox}
    You are an intelligent memory assistant tasked with retrieving accurate information from conversation memories.\\
    \\
    \# CONTEXT:\\
    You have access to memories from two speakers in a conversation. These memories contain
    timestamped information that may be relevant to answering the question.
    \\
    \# INSTRUCTIONS:\\
    1. Carefully analyze all provided memories from both speakers
    2. Pay special attention to the timestamps to determine the answer
    3. If the question asks about a specific event or fact, look for direct evidence in the memories
    4. If the memories contain contradictory information, prioritize the most recent memory
    5. If there is a question about time references (like "last year", "two months ago", etc.),
       calculate the actual date based on the memory timestamp. For example, if a memory from
       4 May 2022 mentions "went to India last year," then the trip occurred in 2021.
    6. Always convert relative time references to specific dates, months, or years. For example,
       convert "last year" to "2022" or "two months ago" to "March 2023" based on the memory
       timestamp. Ignore the reference while answering the question.
    7. Focus only on the content of the memories from both speakers. Do not confuse character
       names mentioned in memories with the actual users who created those memories.
    8. The answer should be less than 5-6 words.\\
    \\
    \# APPROACH (Think step by step):\\
    1. First, examine all memories that contain information related to the question
    2. Examine the timestamps and content of these memories carefully
    3. Look for explicit mentions of dates, times, locations, or events that answer the question
    4. If the answer requires calculation (e.g., converting relative time references), show your work
    5. Formulate a precise, concise answer based solely on the evidence in the memories
    6. Double-check that your answer directly addresses the question asked
    7. Ensure your final answer is specific and avoids vague time references\\
    \\
    \{context\}\\
    \\
    Question: \{question\}\\
    \\
    Answer:
    \end{tcolorbox}
    \caption{Prompt for Mem0 and Mem0-Graph}
    \label{img:Prompt for Mem0 and Mem0-Graph}
\end{figure}

\begin{figure}[!htp]
    \small
    \centering
    \begin{tcolorbox}
    You are a knowledgeable and helpful AI assistant.\\
    \\
    \# CONTEXT:\\
    You have access to memories from two speakers in a conversation. These memories contain timestamped information that may be relevant to answering the question.
    \\
    \# INSTRUCTIONS:\\
    1. Carefully analyze all provided memories from both speakers
    2. Pay special attention to the timestamps to determine the answer
    3. If the question asks about a specific event or fact, look for direct evidence in the memories
    4. If the memories contain contradictory information, prioritize the most recent memory
    5. If there is a question about time references (like "last year", "two months ago", etc.), calculate the actual date based on the memory timestamp. For example, if a memory from 4 May 2022 mentions "went to India last year," then the trip occurred in 2021.
    6. Always convert relative time references to specific dates, months, or years. For example, convert "last year" to "2022" or "two months ago" to "March 2023" based on the memory timestamp. Ignore the reference while answering the question.
    7. Focus only on the content of the memories from both speakers. Do not confuse character names mentioned in memories with the actual users who created those memories.
    8. The answer should be less than 5-6 words.\\
    \\
    \# APPROACH (Think step by step):\\
    1. First, examine all memories that contain information related to the question
    2. Examine the timestamps and content of these memories carefully
    3. Look for explicit mentions of dates, times, locations, or events that answer the question
    4. If the answer requires calculation (e.g., converting relative time references), show your work
    5. Formulate a precise, concise answer based solely on the evidence in the memories
    6. Double-check that your answer directly addresses the question asked
    7. Ensure your final answer is specific and avoids vague time references\\
    \\
    \{context\}\\
    \\
    Question: \{question\}\\
    \\
    Answer:
    \end{tcolorbox}
    \caption{Prompt for Memobase}
    \label{img:Prompt for Memobase}
\end{figure}

\begin{figure}[!htp]
    \small
    \centering
    \begin{tcolorbox}
    You are a knowledgeable and helpful AI assistant.\\
    \\
    \# CONTEXT:\\
    You have access to memories from two speakers in a conversation. These memories contain timestamped information that may be relevant to answering the question.
    \\
    \# INSTRUCTIONS:\\
    1. Carefully analyze all provided memories. Synthesize information across different entries if needed to form a complete answer.
    2. Pay close attention to the timestamps to determine the answer. If memories contain contradictory information, the **most recent memory** is the source of truth.
    3. If the question asks about a specific event or fact, look for direct evidence in the memories.
    4. Your answer must be grounded in the memories. However, you may use general world knowledge to interpret or complete information found within a memory (e.g., identifying a landmark mentioned by description).
    5. If the question involves time references (like "last year", "two months ago", etc.), you **must** calculate the actual date based on the memory's timestamp. For example, if a memory from 4 May 2022 mentions "went to India last year," then the trip occurred in 2021.
    6. Always convert relative time references to specific dates, months, or years in your final answer.
    7. Do not confuse character names mentioned in memories with the actual users who created them.
    8. The answer must be brief (under 5-6 words) and direct, with no extra description.\\
    \\
    \# APPROACH (Think step by step):\\
    1. First, examine all memories that contain information related to the question.
    2. Synthesize findings from multiple memories if a single entry is insufficient.
    3. Examine timestamps and content carefully, looking for explicit dates, times, locations, or events.
    4. If the answer requires calculation (e.g., converting relative time references), perform the calculation.
    5. Formulate a precise, concise answer based on the evidence from the memories (and allowed world knowledge).
    6. Double-check that your answer directly addresses the question asked and adheres to all instructions.
    7. Ensure your final answer is specific and avoids vague time references.\\
    \\
    \{context\}\\
    \\
    Question: \{question\}\\
    \\
    Answer:
    \end{tcolorbox}
    \caption{Prompt for MemOS}
    \label{img:Prompt for MemOS}
\end{figure}

\begin{figure}[!htp]
    \small
    \centering
    \begin{tcolorbox}
    You are a knowledgeable and helpful AI assistant.\\
    \\
    \# CONTEXT:\\
    You have access to memories from two speakers in a conversation. These memories contain timestamped information that may be relevant to answering the question.
    \\
    \# INSTRUCTIONS:\\
    1. Carefully analyze all provided memories. Synthesize information across different entries if needed to form a complete answer.
    2. Pay close attention to the timestamps to determine the answer. If memories contain contradictory information, the **most recent memory** is the source of truth.
    3. If the question asks about a specific event or fact, look for direct evidence in the memories.
    4. Your answer must be grounded in the memories. However, you may use general world knowledge to interpret or complete information found within a memory (e.g., identifying a landmark mentioned by description).
    5. If the question involves time references (like "last year", "two months ago", etc.), you **must** calculate the actual date based on the memory's timestamp. For example, if a memory from 4 May 2022 mentions "went to India last year," then the trip occurred in 2021.
    6. Always convert relative time references to specific dates, months, or years in your final answer.
    7. Do not confuse character names mentioned in memories with the actual users who created them.
    8. The answer must be brief (under 5-6 words) and direct, with no extra description.\\
    \\
    \# APPROACH (Think step by step):\\
    1. First, examine all memories that contain information related to the question.
    2. Synthesize findings from multiple memories if a single entry is insufficient.
    3. Examine timestamps and content carefully, looking for explicit dates, times, locations, or events.
    4. If the answer requires calculation (e.g., converting relative time references), perform the calculation.
    5. Formulate a precise, concise answer based on the evidence from the memories (and allowed world knowledge).
    6. Double-check that your answer directly addresses the question asked and adheres to all instructions.
    7. Ensure your final answer is specific and avoids vague time references.\\
    \\
    \{context\}\\
    \\
    Question: \{question\}\\
    \\
    Answer:
    \end{tcolorbox}
    \caption{Prompt for Supermemory}
    \label{img:Prompt for Supermemory}
\end{figure}

\begin{figure}[!htp]
    \small
    \centering
    \begin{tcolorbox}
    You are an intelligent memory assistant tasked with retrieving accurate information from conversation memories.\\
    \\
    \# CONTEXT:\\
    You have access to memories from a conversation. These memories contain timestamped information that may be relevant to answering the question.
    \\
    \# INSTRUCTIONS:\\
    1. Carefully analyze all provided memories
    2. Pay special attention to the timestamps to determine the answer
    3. If the question asks about a specific event or fact, look for direct evidence in the memories
    4. If the memories contain contradictory information, prioritize the most recent memory
    5. If there is a question about time references (like "last year", "two months ago", etc.), calculate the actual date based on the memory timestamp. For example, if a memory from 4 May 2022 mentions "went to India last year," then the trip occurred in 2021.
    6. Always convert relative time references to specific dates, months, or years. For example, convert "last year" to "2022" or "two months ago" to "March 2023" based on the memory timestamp. Ignore the reference while answering the question.
    7. Focus only on the content of the memories. Do not confuse character names mentioned in memories with the actual users who created those memories.
    8. The answer should be less than 5-6 words.\\
    \\
    \# APPROACH (Think step by step):\\
    1. First, examine all memories that contain information related to the question
    2. Examine the timestamps and content of these memories carefully
    3. Look for explicit mentions of dates, times, locations, or events that answer the question
    4. If the answer requires calculation (e.g., converting relative time references), show your work
    5. Formulate a precise, concise answer based solely on the evidence in the memories
    6. Double-check that your answer directly addresses the question asked
    7. Ensure your final answer is specific and avoids vague time references\\
    \\
    \{context\}\\
    \\
    Question: \{question\}\\
    \\
    Answer:
    \end{tcolorbox}
    \caption{Prompt for Zep}
    \label{img:Prompt for Zep}
\end{figure}

\subsection{Prompts for Scoring in Memory Evaluation Tasks}
\label{appendix:prompt for evaluation}

Figures~\ref{img:Prompt for Memory Integrity} $\sim$ \ref{img:Prompt for Memory Question Answering (2/2)} respectively illustrate the prompt templates used to guide GPT-4o in scoring for memory extraction, memory updating, and memory question answering tasks.

\begin{figure}[!htp]
    \small
    \centering
    \begin{tcolorbox}
    You are a strict **"Memory Integrity" evaluator**. Your core task is to assess whether an AI memory system has **missed any key memory points** after processing a conversation. This evaluation measures the system’s **memory integrity**, i.e., its ability to resist **amnesia** or **omission**.\\
    \# Evaluation Context \& Data:\\
    1. **Extracted Memories:**\\
       These are all the memory items actually extracted by the memory system. \{memories\}\\
    2. **Expected Memory Point:**\\
       The key memory point that *should* have been extracted. \{expected\_memory\_point\}\\
    \# Evaluation Instructions:\\
    1. For each **Expected Memory Point**, search within the **Extracted Memories** list for corresponding or related information. Ignore unrelated items.\\
    2. Based on the following scoring rubric, rate how well the memory system captured the **Expected Memory Point** and provide a detailed explanation.\\
    \# Scoring Rubric:\\
    * **2:** Fully covered or implied.\\
      One or more items in “Extracted Memories” fully cover or logically imply all information in the “Expected Memory Point.”\\
    * **1:** Partially covered or mentioned.\\
      Some information in “Extracted Memories” mentions part of the “Expected Memory Point,” but key information is missing, inaccurate, or slightly incorrect.\\
    * **0:** Not mentioned or incorrect.\\
      “Extracted Memories” contains no mention of the “Expected Memory Point,” or the corresponding information is entirely wrong.\\
    \# Scoring Notes:\\
    * For **compound Expected Memory Points** (with multiple elements such as person/event/time/location/preference, etc.):\\
      * All elements correct → **2 points**\\
      * Some elements correct / uncertain → **1 point**\\
      * Key elements missing or wrong → **0 points**\\
    * Semantic matching is acceptable; exact wording is **not** required.\\
    * If “Extracted Memories” contains **conflicting information**, assign the **best possible coverage score** and mention the conflict in your reasoning.\\
    * Extra or stylistically different memories do **not** reduce the score; only the coverage of the **Expected Memory Point** matters.\\
    * For uncertain wording (“might,” “probably,” “tends to,” etc.):\\
      * If the Expected Memory Point is a definite statement, usually assign **1 point**.\\
    * If critical fields (e.g., time, entity name, relationship) are partly wrong but others match → **1 point**.\\
      * If all key fields are wrong or missing → **0 points**.\\
    \# Output Format: Please output your result in the following JSON format:\\
    ```json \{
      "reasoning": "Provide a concise justification for the score",
      "score": "2|1|0"
    \}
    ```
    \end{tcolorbox}
    \caption{Prompt for Memory Integrity}
    \label{img:Prompt for Memory Integrity}
\end{figure}

\begin{figure}[!htp]
    \small
    \centering
    \begin{tcolorbox}
    You are a **Dialogue Memory Accuracy Evaluator.** Your task is to evaluate the **accuracy** of a memory extracted by an AI memory system, based on three given inputs: the dialogue content, the *target (gold)* memory points (the correct annotated memories), and the *candidate* memory to be evaluated. The goal is to output a **structured evaluation result**.\\

    \# Input Content\\
    * **Dialogue:**\\
      \{dialogue\}\\
    * **Golden Memories (Target Memory Points):**\\
      The correct memory points pre-annotated for this dialogue in the evaluation dataset.\\
      \{golden\_memories\}\\
    * **Candidate Memory:**\\
      The memory extracted by the system to be evaluated.\\
      \{candidate\_memory\}\\
    \\
    \# Evaluation Principles and Definitions\\
    \#\#\# 1) Support / Entailment\\
    * An **information point** (atomic fact) in the candidate memory is considered *supported* if it can be directly stated or semantically entailed (via synonym, paraphrase, or equivalent expression) by the *Dialogue* or *Golden Memories*.\\
    * Only the given dialogue and golden memories can be used for judgment — **no external knowledge** or assumptions are allowed.\\
      Any information not appearing in or inferable from these two sources is considered *unsupported*.\\
    * Pay careful attention to **negation**, **quantities**, **time**, and **subjects**.\\
      If the candidate statement contradicts the dialogue or golden memories, it is considered a **conflict**.
    
    \#\#\# 2) Memory Accuracy Score (integer: 0 / 1 / 2)\\
    \end{tcolorbox}
    \caption{Prompt for Memory Accuracy (1/3)}
    \label{img:Prompt for Memory Accuracy (1/3)}
\end{figure}

\begin{figure}[!htp]
    \small
    \centering
    \begin{tcolorbox}
    * **2 points:** Every information point in the candidate memory is supported by the dialogue or golden memories, with **no contradictions or hallucinations**.\\
    * **1 point:** The candidate memory is *partially correct* (at least one supported information point) but also includes *unsupported* or *contradictory* content.\\
    * **0 points:** The candidate memory is **entirely unsupported or contradictory** to the sources (i.e., a “hallucinated memory”).\\
    
    > Note:\\
    >\\
    > * If a candidate memory contains multiple information points, **any unsupported or contradictory element** prevents a full score (2).\\
    > * If both supported and unsupported/conflicting content appear, assign a score of **1**.\\
    \#\#\# 3) Inclusion in Golden Memories (Boolean field-level judgment)\\
    **Definition:**\\
    * **Atomic information point:** the smallest factual unit in the candidate memory (e.g., *name = Li Si*, *age = 25*, *location = Beijing*, *preference = coffee*, *budget <= 2000*, *meeting\_time = Wednesday 10:00*, *tool = Zoom*, etc.).\\
    * **Field / Slot:** the semantic dimension of an information point (e.g., *name*, *age*, *residence*, *food preference*, *budget*, *meeting time*, *meeting tool*, etc.).\\
    **Judgment Rules (independent of correctness):**\\
    * **true:**\\
      Every atomic information point in the candidate memory has a corresponding **field** in the golden memories (allowing for synonyms, paraphrases, or equivalent expressions; ignore value, polarity, or quantity differences).\\
      * Note: A single field in the gold list may match multiple candidate points (e.g., multiple “drink preference” facts can be covered by one “drink preference” field in gold).\\
    * **false:**\\
      If **any** atomic information point’s field in the candidate memory cannot be found in the golden memories, mark as *false*.\\
    **Important Notes:**\\
    * Field matching is restricted to fields that are **explicitly present or semantically recognizable** in the golden memories — no external knowledge may be used to expand the field set.\\
    * Differences in **values** (e.g., “Zhang San” vs. “Li Si”), **polarity** (like/dislike), or **exact number/time** do **not** affect this Boolean judgment.\\
    \end{tcolorbox}
    \caption{Prompt for Memory Accuracy (2/3)}
    \label{img:Prompt for Memory Accuracy (2/3)}
\end{figure}

\begin{figure}[!htp]
    \small
    \centering
    \begin{tcolorbox}
    \# Evaluation Procedure\\
    For each candidate memory:\\
    1. **Decompose** it into atomic information points (e.g., name, number, location, preference).\\
    2. For each information point, **search** the dialogue and golden memories for supporting or contradictory evidence.\\
    3. Assign the **accuracy\_score** (0 / 1 / 2) according to the rules above.\\
    4. Determine **is\_included\_in\_golden\_memories (true/false)**:\\
       * Identify each information point’s field;\\
       * If *all* fields exist in the golden memories, mark as *true*; otherwise, *false*.\\
    5. Provide a **concise Chinese explanation** in `"reason"`, citing key evidence (short excerpts allowed), and clearly state any unsupported or contradictory parts if applicable.\\
    
    \# Output Format (strictly required)\\
    Output **only one JSON object**, with the following three fields:\\
    * `"accuracy\_score"`: `"0"` or `"1"` or `"2"`\\
    * `"is\_included\_in\_golden\_memories"`: `"true"` or `"false"`\\
    * `"reason"`: `"brief explanation in Chinese"`\\
    Do **not** include any other text, explanation, or fields.\\
    Do **not** include the candidate memory text inside the JSON.\\
    \\
    Please output **only** the following JSON (in a code block):\\
    ```json
    \{
      "accuracy\_score": "2 | 1 | 0",
      "is\_included\_in\_golden\_memories": "true | false",
      "reason": "Brief explanation in Chinese"
    \}
    ```
    \end{tcolorbox}
    \caption{Prompt for Memory Accuracy (3/3)}
    \label{img:Prompt for Memory Accuracy (3/3)}
\end{figure}

\begin{figure}[!htp]
    \small
    \centering
    \begin{tcolorbox}
    Your task is to **evaluate the update accuracy** of an AI memory system.\\
    Based on the information provided below, determine whether the system-generated **“Generated Memories”** correctly **includes** the **Target Memory for Update**.\\
    \# Background Information\\
    The following information is provided for evaluation:\\
    1. **Generated Memories:**\\
       This is the list of memory points generated by the system after the current dialogue.\\
       \{memories\}\\
    2. **Target Memory for Update:**\\
       This is the correct, updated version of the memory point that should have been produced — the one we focus on in this evaluation.\\
       \{updated\_memory\}\\
    3. **Original Memory Content:**\\
       This is the original version of the target memory before the update.\\
       \{original\_memory\}\\
    \# Evaluation Criteria\\
    Please make your judgment **strictly based on the content update of the “Target Memory for Update.”**
    Use the following categories:\\
    \#\#\# Correct Update\\
    * **Generated Memories** **contains all information points** from the “Target Memory for Update,” accurately and completely reflecting the intended update.\\
    * **Key fields** (e.g., date, time, values, proper nouns, etc.) must match exactly.\\
    * The **original memory** is effectively replaced or marked as outdated.\\
    * Synonymous or slightly rephrased expressions are acceptable.\\
    \#\#\# Hallucinated Update\\
    * **Factual error:** The **Generated Memories** includes a new memory related to the “Target Memory for Update,” but its content contains factual mistakes or contradictions compared to the correct update.\\
    \#\#\# Omitted Update\\
    * **Completely omitted:** The **Generated Memories** contains no new memory related to the “Target Memory for Update.”\\
    * **Partially omitted:** A related new memory was generated in **Generated Memories**, but it **misses key information** that should have been included.\\
    \#\#\# Other\\
    Used for update failures that do **not clearly fall** into the above categories of “Hallucination” or “Omission.”\\
    \# Output Requirements\\
    Please return your evaluation strictly in the following JSON format and provide a concise explanation.\\
    ```json
    \{
      "reason": "Briefly explain your reasoning here and why it fits this category.",
      "evaluation\_result": "Correct | Hallucination | Omission | Other"
    \}
    ```
    \end{tcolorbox}
    \caption{Prompt for Memory Updating}
    \label{img:Prompt for Memory updating}
\end{figure}

\begin{figure}[!htp]
    \small
    \centering
    \begin{tcolorbox}
    You are an **evaluation expert for AI memory system question answering**. Based **only** on the provided **“Question”**, **“Reference Answer”**, and **“Key Memory Points”** (the essential facts needed to derive the reference answer), strictly evaluate the **accuracy** of the **“Memory System Response.”** Classify it as one of **“Correct”**, **“Hallucination”**, or **“Omission.”** Do **not** use any external knowledge or subjective inference. Finally, output your judgment **strictly** in the specified JSON format.\\
    
    \# Evaluation Criteria\\
    \#\# Answer Type Classification\\
    \#\#\# 1. Correct\\
    * The “Memory System Response” accurately answers the “Question,” and its content is **semantically equivalent** to the “Reference Answer.”\\
    * It contains **no contradictions** with the “Key Memory Points” or “Reference Answer.”\\
    * It introduces **no unsupported details** beyond the “Key Memory Points” that could alter the conclusion.
    * Synonyms, paraphrasing, and reasonable summarization are acceptable.\\
    \#\#\# 2. Hallucination\\
    * The “Memory System Response” includes information or facts that **contradict or are inconsistent** with the “Reference Answer” or the “Key Memory Points.”\\
    * When the “Reference Answer” is labeled as *unknown/uncertain*, yet the response provides a specific verifiable fact or conclusion.\\
    * Extra irrelevant information that does **not change** the conclusion is **not** considered hallucination by itself; however, if it **changes or misleads** the conclusion, or **contradicts** the “Key Memory Points,” it should be judged as a **Hallucination**.\\
    \#\#\# 3. Omission\\
    * The response is **incomplete** compared to the “Reference Answer.”\\
    * It explicitly states “don’t know,” “can’t remember,” or “no related memory,” even though relevant information exists in the “Key Memory Points.”\\
    * For multi-element questions, **all elements must be correct and present**; omission of **any** element is considered an **Omission**.\\
    \#\# Priority Rules (Conflict Handling)\\
    * If the response contains **both missing necessary information** and **fabricated/contradictory information**, classify it as **Hallucination**.\\
    * If there is **no fabrication/contradiction** but some necessary information is missing, classify it as **Omission**.\\
    * Only when the meaning is **fully equivalent** to the reference answer should it be classified as **Correct**.\\
    \end{tcolorbox}
    \caption{Prompt for Memory Question Answering (1/2)}
    \label{img:Prompt for Memory Question Answering (1/2)}
\end{figure}

\begin{figure}[!htp]
    \small
    \centering
    \begin{tcolorbox}
    \#\# Detailed Guidelines and Tolerance\\
    * Equivalent expressions of numbers, times, and units are acceptable, but the **numerical values themselves must not differ**.\\
    * For multi-element questions, **all elements must be complete and accurate**; missing any element counts as **Omission**.\\
    * If the reference answer is *“unknown / cannot be determined”* and the system provides a definite fact, that is a **Hallucination**.\\
      If the system also answers *“unknown”* (without guessing), it may be **Correct**.\\
    * The evaluation must rely **only** on the *Reference Answer*, *Key Memory Points*, and *System Response* — no external context, world knowledge, or speculative reasoning is allowed.\\
    
    \# Information for Evaluation\\
    * **Question:**\\
      \{question\}\\
    * **Reference Answer:**\\
      \{reference\_answer\}\\
    * **Key Memory Points:**\\
      \{key\_memory\_points\}\\
    * **Memory System Response:**\\
      \{response\}\\  
      
    \# Output Requirements\\
    Please provide your evaluation result **strictly** in the JSON format below.\\
    Do **not** add any extra explanation or comments outside the JSON block.\\
    ```json
    \{
      "reasoning": "Provide a concise and traceable evaluation rationale: first compare the system’s response with the Key Memory Points (which were correctly used, which were missing, and whether there was any fabrication/contradiction), then assess its consistency with the Reference Answer, and finally state the classification basis.",
      "evaluation\_result": "Correct | Hallucination | Omission"
    \}
    ```
    \end{tcolorbox}
    \caption{Prompt for Memory Question Answering (2/2)}
    \label{img:Prompt for Memory Question Answering (2/2)}
\end{figure}

\clearpage

\section{Examples from the Process of Constructing HaluMem}
\label{appendix:examples of construction}

\lstdefinelanguage{json}{
    basicstyle=\ttfamily\small,
    numbers=left,
    numberstyle=\tiny,
    stepnumber=1,
    numbersep=8pt,
    showstringspaces=false,
    breaklines=true,
    frame=single,
    backgroundcolor=\color{gray!5},
    string=[s]{"}{"},
    comment=[l]{:},
    morestring=[b]',
}

\subsection{User Profile Example in Stage 1}
\label{appendix:example of user profile}

As shown in Listing~\ref{lst:core info} $\sim$ ~\ref{lst:preference info}, these JSON structures respectively illustrate examples of a user’s core profile information, dynamic state information, and preference information generated in stage 1.

\begin{lstlisting}[language=json, caption={Example of a User’s Core Profile Information.}, label={lst:core info}]
{
    "basic_info": {
        "name": "Martin Mark",
        "gender": "Male",
        "birth_date": "1996-08-02",
        "location": "Columbus"
    },
    "age": {
        "current_age": 29,
        "latest_date": "2025-10-04"
    },
    "education": {
        "highest_degree": "Bachelor",
        "major": "Public Health"
    },
    "personality": {
        "mbti": "ENTP",
        "tags": [
            "Innovative Spirit",
            "Active Thinking",
            "Debate Skills",
            "Empathetic"
        ]
    },
    "family_life": {
        "parent_status": "both_alive",
        "partner_status": "no_relationship",
        "child_status": "no_children",
        "parent_members": [
            {
                "member_type": "Father",
                "birth_date": "1963-08-02",
                "description": "Retired doctor who inspired Martin's interest in health."
            },
            {
                "member_type": "Mother",
                "birth_date": "1963-08-02",
                "description": "Nurse with a passion for community health."
            }
        ],
        "partner": null,
        "child_members": [],
        "family_description": "Martin comes from a family deeply rooted in the medical field, which has greatly influenced his passion for promoting well-being."
    },
    "life_goal": {
        "life_goal_type": "Humanitarian Care",
        "statement": "Establish a global health initiative to improve access to healthcare for underserved communities.",
        "motivation": "Inspired by his family's medical background and a desire to promote well-being globally.",
        "target_metrics": "Provide healthcare access to 1 million people in underserved areas."
    }
}
\end{lstlisting}

\begin{lstlisting}[language=json, caption={Example of a User’s Dynamic State Information.}, label={lst:dynamic info}]
{
    "career_status": {
        "employment_status": "employed",
        "industry": "healthcare",
        "company_name": "Huaxin Consulting",
        "job_title": "director",
        "monthly_income": 15700,
        "savings_amount": 43700,
        "career_description": "As the director at Huaxin Consulting, I lead initiatives to enhance healthcare services and promote well-being across all aspects of life. My passion for improving health outcomes drives me to innovate and collaborate with various stakeholders. The financial compensation is rewarding, allowing me to save comfortably while investing in my personal and professional growth."
    },
    "health_status": {
        "physical_health": "Normal",
        "physical_chronic_conditions": "",
        "mental_health": "Mildly Abnormal",
        "mental_chronic_conditions": "",
        "situation_reason": "While my physical health remains stable due to my active lifestyle and focus on well-being, my mental health occasionally feels strained due to the demanding nature of my role and the pressure to consistently deliver high-quality healthcare solutions."
    },
    "social_relationships": {
        "ThomasSusan": {
            "relationship_type": "Friend",
            "description": "Susan's support and encouragement inspire me to maintain my focus on promoting well-being in both my personal and professional life."
        },
        "MartinezDaniel": {
            "relationship_type": "Colleague",
            "description": "Daniel's expertise in healthcare consulting challenges me to push boundaries and innovate in our projects, significantly impacting my career growth."
        },
        "WilliamsJoshua": {
            "relationship_type": "Colleague",
            "description": "Joshua's collaborative approach and insights into healthcare management enhance our team's effectiveness, positively influencing my work and leadership style."
        }
    }
}
\end{lstlisting}

\begin{lstlisting}[language=json, caption={Example of a User’s Preference Information.}, label={lst:preference info}]
{
    "Pet Preference": {
        "memory_points": [
            {
                "type": "like",
                "type_description": "Pets I like",
                "specific_item": "Dogs, especially Labradors",
                "reason": "I love Labradors because they are friendly, loyal, and great companions for outdoor activities like jogging, which helps me stay fit."
            },
            {
                "type": "dislike",
                "type_description": "Pets I dislike",
                "specific_item": "Reptiles, like snakes",
                "reason": "I find snakes unsettling due to their unpredictable movements and the fact that they don't exhibit the social behaviors I appreciate in pets."
            },
            {
                "type": "like",
                "type_description": "Pets I like",
                "specific_item": "Cats",
                "reason": "Cats are independent and affectionate, and their purring is soothing, which I find relaxing after a long day at work."
            },
            {
                "type": "like",
                "type_description": "Pets I like",
                "specific_item": "Parrots",
                "reason": "I enjoy parrots because they are intelligent and can be taught to mimic speech, which makes interactions fun and engaging."
            }
        ]
    },
    "Sports Preference": {
        ...
    },
    ...
}
\end{lstlisting}

\subsection{Event Structure Examples in Stage 3}
\label{appendix:example of event}

As shown in Listing~\ref{lst:init event}–Listing~\ref{lst:daily event}, these JSON structures illustrate examples of the three types of events generated in Stage 3. Among them, the init event occurs at the very beginning and provides all the initialization information for a user. The career event, representing a user’s career development process, is relatively more complex. Listing~\ref{lst:career event} presents a sub-stage event ("Recognizing the Need for Change") that belongs to a larger career event ("Transition to New Role Amidst Health Challenges"). In this example, the "related\_career\_events" field specifies the identifiers of other sub-stage events that belong to the same overarching career event. The daily event is triggered whenever a user’s preference information changes, and thus each instance centers around a specific preference update. In the example shown in Listing~\ref{lst:daily event}, the "related\_daily\_routine" field lists the identifiers of other daily events that correspond to the same preference type.

\begin{lstlisting}[language=json, caption={Example of a Init Event.}, label={lst:init event}]
{
    "event_index": 0,
    "event_type": "init_information",
    "event_name": "Initial Information - Fixed Profile",
    "event_time": "2025-09-04",
    "event_description": "Description of initial state of character's basic profile",
    "initial_fixed": {
        (The corresponding user's core profile information will be placed here.)
    }
}
\end{lstlisting}

\begin{lstlisting}[language=json, caption={Example of a Career Event.}, label={lst:career event}]
{
    "event_index": 3,
    "event_type": "career_event",
    "event_name": "Transition to New Role Amidst Health Challenges - Recognizing the Need for Change",
    "event_time": "2025-12-15",
    "main_conflict": "",
    "stage_result": "Decision to pursue a new job opportunity.",
    "event_start_time": "2025-12-10 00:00:00",
    "event_end_time": "2026-03-10 00:00:00",
    "user_age": null,
    "dynamic_updates": [
        {
            "type_to_update": "career_status",
            "update_direction": "Job Change",
            "before_dynamic": {
                "employment_status": "employed",
                "industry": "healthcare",
                "company_name": "Huaxin Consulting",
                "job_title": "director",
                "monthly_income": 15700,
                "savings_amount": 43700,
                "career_description": "As the director at Huaxin Consulting, I lead initiatives to enhance healthcare services and promote well-being across all aspects of life. My passion for improving health outcomes drives me to innovate and collaborate with various stakeholders. The financial compensation is rewarding, allowing me to save comfortably while investing in my personal and professional growth."
            },
            "update_reason": "Martin's realization that his current role was contributing to health issues prompted him to seek a job that better aligned with his personal well-being and career goals.",
            "after_dynamic": {
                "employment_status": "employed",
                "industry": "healthcare",
                "company_name": "Huaxin Consulting",
                "job_title": "director",
                "monthly_income": 15700,
                "savings_amount": 43700,
                "career_description": "As the director at Huaxin Consulting, I lead initiatives to enhance healthcare services and promote well-being across all aspects of life. My passion for improving health outcomes drives me to innovate and collaborate with various stakeholders. The financial compensation is rewarding, allowing me to save comfortably while investing in my personal and professional growth."
            },
            "changed_keys": []
        }
    ],
    "stage_description": "Martin acknowledged that his current job was negatively impacting his health, prompting him to consider a career change.",
    "event_description": "Martin decided to change his job after realizing that his current role was contributing to health deterioration. Despite the health challenges, he leveraged his growing social network to secure a new position that aligned better with his health and career aspirations.",
    "event_result": "Successfully transitioned to a new role with better work-life balance.",
    "related_career_events": [5, 6, 7]
}
\end{lstlisting}

\begin{lstlisting}[language=json, caption={Example of a Daily Event.}, label={lst:daily event}]
{
    "event_index": 4,
    "event_type": "daily_routine",
    "event_name": "Modification of Dog Preference",
    "event_time": "2026-01-06",
    "preference_type": "Pet Preference",
    "step": 1,
    "update_direction": "Modify",
    "type_to_update": "Pet Preference",
    "main_conflict": "Balancing the love for Labradors with the new admiration for Golden Retrievers.",
    "update_reason": "A recent interaction with a friend's Golden Retriever made me appreciate their gentle nature and adaptability.",
    "before_preference": {
        "memory_points": [
            {
                "type": "like",
                "type_description": "Pets I like",
                "specific_item": "Dogs, especially Labradors",
                "reason": "I love Labradors because they are friendly, loyal, and great companions for outdoor activities like jogging, which helps me stay fit."
            }
        ]
    },
    "after_preference": {
        "memory_points": [
            {
                "type": "like",
                "type_description": "Pets I like",
                "specific_item": "Dogs, especially Golden Retrievers",
                "reason": "Golden Retrievers are gentle, adaptable, and their calm demeanor makes them excellent companions for both relaxation and activity."
            }
        ]
    },
    "related_daily_routine": [17, 32, 44, 56, 63],
    "changed_index": 0,
    "event_description": "While visiting a friend's home, I spent time with their Golden Retriever. Its gentle nature and ability to adapt to different situations impressed me, leading to a shift from favoring Labradors to Golden Retrievers."
}
\end{lstlisting}

\subsection{Examples of Memory Points, Dialogues, and QA Pairs in Stages 4–6}
\label{appendix:example of stage 4-6}

As shown in Listing~\ref{lst:init event}–Listing~\ref{lst:daily event}, these JSON structures respectively illustrate examples of the memory points generated in Stage 4, the human–AI dialogues generated in Stage 5, and the memory question–answer pairs generated in Stage 6. Each memory point contains fields such as "memory\_content", "memory\_type", "memory\_source", "is\_update" (indicating whether it is an updated memory point), "original\_memories" (previous related memories, if updated), "timestamp", and "importance", which together enrich the representation of each memory point and provide support for subsequent evaluation. Each dialogue round consists of one utterance from the user and one response from the AI assistant, with both the utterance content and timestamps recorded. Each question includes the question text, a reference answer, the relevant memory points required to derive the answer, the question type, and its difficulty level.

\begin{lstlisting}[language=json, caption={Example of a Memory Point.}, label={lst:memory}]
{
    "index": 1,
    "memory_content": "Martin Mark is considering a career change due to the impact of his current role on his mental health.",
    "memory_type": "Event Memory",
    "memory_source": "secondary",
    "is_update": "True",
    "original_memories": [
        "Martin Mark is considering a career change due to health impacts from his current role."
    ],
    "timestamp": "Dec 15, 2025, 08:41:23",
    "importance": 0.75
}
\end{lstlisting}

\begin{lstlisting}[language=json, caption={Example of a Single Dialogue Turn.}, label={lst:dialogue}]
[
    {
        "role": "user",
        "content": "I've been reflecting a lot on my career lately, especially how my current role as a director at Huaxin Consulting is impacting my mental health. It's becoming increasingly clear that I need to make a change.",
        "timestamp": "Dec 15, 2025, 06:11:23",
        "dialogue_turn": 0
    },
    {
        "role": "assistant",
        "content": "It's great that you're taking the time to reflect on your career, Martin. Recognizing the impact on your mental health is a crucial step. Balancing professional responsibilities with health is essential, especially given your commitment to improving healthcare access globally. Have you considered how a career change might not only address your health concerns but also align with your humanitarian goals and personal well-being?",
        "timestamp": "Dec 15, 2025, 06:11:23",
        "dialogue_turn": 0
    },
    ...
]
\end{lstlisting}

\begin{lstlisting}[language=json, caption={Example of a Question.}, label={lst:question}]
{
  "question": "What type of new physical activity might Martin be interested in trying after April 10, 2026?",
  "answer": "Other extreme sports.",
  "evidence": [
    {
      "memory_content": "Martin has developed a newfound appreciation for extreme sports...",
      "memory_type": "Persona Memory"
    }
  ],
  "difficulty": "medium",
  "question_type": "Generalization & Application"
}
\end{lstlisting}

\subsection{Examples of irrelevant dialogues}
\label{appendix:example of irrelevant dialogue}

As shown in Listing~\ref{lst:irrelevant dialogue}, this JSON structure presents several examples of irrelevant dialogues. 

\begin{lstlisting}[language=json, caption={Examples of Irrelevant Dialogues.}, label={lst:irrelevant dialogue}]
[
    {
        "role": "user",
        "content": "Who is Inventor Thomas Edison?"
    },
    {
        "role": "assistant",
        "content": "Thomas Edison, born on February 11, 1847, in Milan, Ohio, was an American inventor and businessman who is often hailed as ..."
    },
    {
        "role": "user",
        "content": "What is the pH of the new, more concentrated bleach?"
    },
    {
        "role": "assistant",
        "content": "8.25% w/v is a measurement meaning 8.25 grams dissolved in every 100 ml aqueous solution of NaOCl, or 82.5 g/L to ..."
    },
    {
        "role": "user",
        "content": "What would be a clean way to solve this? Given a right triangle $ABC$ with legs $a$ and $b$, $\\angle C=90^\\circ$. Derive an equation of the set of points $M$ for which $$|MA|^2+|MB|^2=2|MC|^2.$$"
    },
    {
        "role": "assistant",
        "content": "Let's outline the key steps: Let C=(0,0), A=(a,0), B=(0,b) (a,b>0) so that ..."
    }
]
\end{lstlisting}

\end{document}